# Automatic detection of microsleep episodes with deep learning

**Alexander Malafeev[1,2], Anneke Hertig-Godeschalk[3], David R. Schreier[3], Jelena Skorucak[1,2], Johannes Mathis[3#], Peter Achermann[1, 2, 4, 5#]**

[1]Institute of Pharmacology and Toxicology, University of Zurich, Zurich, Switzerland

[2]Neuroscience Center Zurich, University of Zurich and ETH Zurich, Zurich, Switzerland

[3]Department of Neurology, Inselspital, Bern University Hospital, University of Bern, Bern, Switzerland

[4]The KEY Institute for Brain-Mind Research, Department of Psychiatry, Psychotherapy and Psychosomatics, University Hospital of Psychiatry, Zurich, Switzerland

[5]Sleep and Health, University of Zurich, Zurich, Switzerland

# shared last authorship

**Correspondence:**
Peter Achermann

The KEY Institute for Brain   Mind Research

Hegibachstrasse 30

CH-8032 Zurich

Switzerland

peter.achermann@uzh.ch

**Keywords: microsleep episodes, excessive daytime sleepiness, drowsiness, deep learning, machine learning**

**Abstract**

Brief fragments of sleep shorter than 15 s are defined as microsleep episodes (MSEs), often subjectively perceived as sleepiness. Their main characteristic is a slowing in frequency in the electroencephalogram (EEG), similar to stage N1 sleep according to standard criteria. The maintenance of wakefulness test (MWT) is often used in a clinical setting to assess vigilance. Scoring of the MWT in most sleep-wake centers is limited to classical definition of sleep (30-s epochs), and MSEs are mostly not considered in the absence of established scoring criteria defining MSEs but also because of the laborious work. We aimed for automatic detection of MSEs with machine learning, i.e. with deep learning based on raw EEG and EOG data as input. We analyzed MWT data of 76 patients. Experts visually scored wakefulness, and according to recently developed scoring criteria MSEs, microsleep episode candidates (MSEc), and episodes of drowsiness (ED). We implemented segmentation algorithms based on convolutional neural networks (CNNs) and a combination of a CNN with a long-short term memory (LSTM) network. A LSTM network is a type of a recurrent neural network which has a memory for past events and takes them into account. Data of 53 patients were used for training of the classifiers, 12 for validation and 11 for testing. Our algorithms showed a good performance close to human experts. The detection was very good for wakefulness and MSEs and poor for MSEc and ED, similar to the low inter-expert reliability for these borderline segments. We performed a visualization of the internal representation of the data by the artificial neuronal network performing best using t-distributed stochastic neighbor embedding (t-SNE). Visualization revealed that MSEs and wakefulness were mostly separable, though not entirely, and MSEc and ED largely intersected with the two main classes. We provide a proof of principle that it is feasible to reliably detect MSEs with deep neuronal networks based on raw EEG and EOG data with a performance close to that of human experts. Code of algorithms ( https://github.com/alexander-malafeev/microsleep-detection ) and data ( https://zenodo.org/record/3251716 ) are available.

# 1    INTRODUCTION

Excessive daytime sleepiness (EDS) is a common complaint of many patients (1-4) and also reported by the general population when sleep is chronically curtailed. Accurate diagnosis of the underlying disorders often requires objective evaluation of nocturnal sleep and daytime sleepiness in these patients. State of the art methods to evaluate sleepiness are the Multiple Sleep Latency Test (MSLT) (5) and the Maintenance of Wakefulness Test (MWT) (6).

Microsleep episodes (MSEs) are considered to be an objective sign of excessive daytime sleepiness (EDS) (7). The MWT is the primarily used test to quantify the ability to maintain wakefulness despite the presence of increased sleep pressure subjectively perceived as EDS.

In most of the studies, the latency to sleep stage N1 or any other stages of sleep is used as a definition for objective sleepiness (8-10). However, it is well accepted that signs of sleepiness appear much earlier, not only in the EEG but also in behavioral changes and performance lapses.

Therefore, more sensitive and systematic, but still practically useful definitions of objective sleepiness are needed. The recently developed Bern continuous and high-resolution wake-sleep (BERN) scoring criteria for assessing the wake-sleep transition zone represent such an approach (7). The criteria were developed for visual scoring of MSEs as short as 1 s, which is time consuming. Moreover, no generally accepted scoring criteria exist so far. Thus, tools for automated analysis of such data would be very useful for both clinicians and researchers in order to reduce the workload and the subjectivity of scoring.

In a study subsequent to the development of the BERN scoring criteria, we developed algorithms for machine learning based automatic detection MSEs using manually engineered features mainly derived from spectral information of the electroencephalogram (EEG) (11).

Another interesting approach was taken by authors of the Vigilance Algorithm Leipzig (VIGALL) (12). They established scoring criteria for 7 vigilance stages (1-s resolution; from fully awake to sleep) and developed an algorithm for the automatic scoring of these stages.

The aim of this work was to implement a deep learning approach using raw data as input. We think that such an algorithm resembles human scoring, which is mainly based on visual pattern recognition. It has also been shown that deep learning methods perform better than classical machine learning (ML) methods on various types of data (13), including EEG data (14-18). Automatic sleep classification has been extensively developed mainly due to the advantages in machine learning, and especially in deep learning (15-20).

## 1.1   Our contribution

We developed several artificial neural networks, which work with raw data as input and compared their performance with the inter-rater agreement of two experts. Note that inter-rater agreement was computed only for five recordings, which were scored by two different experts from the same sleep center. It is also important to note that the selection of the recordings for double scoring was not totally random: only recordings containing MSEs were randomly selected for double scoring. Our networks showed similar agreement to a human expert as the inter-rater agreement between two human experts. We also performed visualization of the hidden representation of the data by one of the networks, the one performing best, using a t-distributed stochastic neighbor embedding (t-SNE) method (21). The code for this paper is available at https://github.com/alexander-malafeev/microsleep-detection .

## 2    MATERIAL AND METHODS

### 2.1    Data

MWT data from 76 patients with EDS recorded at approximately 15:00 were retrospectively analyzed. The suspected diagnosis widely varied between patients (Table 1) and included sleep apnea, narcolepsy, idiopathic hypersomnia, non-organic hypersomnia, and insomnia (11). Patients were not stratified into subgroups because only few patients were available with a certain suspected diagnosis due to their low prevalence. Among other data, recordings included EEG, electrooculogram (EOG), and video recordings of the face (7, 11). Electrophysiological signals were sampled at 200 Hz (band pass filter 0.3-70 Hz; 50 Hz notch filter; RemLogic™ (Embla Systems LLC)) and exported in the European data format (EDF) for post processing.

MSEs were visually scored by a sleep expert using both occipital EEG derivations referenced to contralateral mastoid electrodes (i.e. channels O1M2 and O2M1), two EOG channels, both referenced to the left mastoid electrode (i.e. channels E1M1 and E2M1), and video recordings of the face. Video recordings were not used for automatic detection algorithm, only EEG and EOG data were considered. MSEs were defined as 1-15 s in duration with a clear slowing in the EEG resulting in a theta dominance resembling non-rapid eye movement (NREM) sleep stage 1 (N1), while at least an 80% eye closure was observed in the video recording. MSEs were typically preceded by slow eye movements, visible in the EOG. Apart from clear wakefulness and MSEs, two poorly defined EEG patterns were categorized as microsleep episode candidates (MSEc; not fulfilling all of the criteria for a MSE, e.g. eyes were closed less than 80%) and episodes of drowsiness (ED; even more vague, not clear wake or MSE or MSEc) (7). Approximately 2/3 of the recordings were checked by another expert and differences were resolved by discussion. The beginning and the end of each episode was marked continuously, i.e. with the resolution of the recording (1/200 s).

Each MWT lasted 40 min and was supposed to be terminated earlier if consolidated sleep occurred (7, 11). However, if the technician missed terminating the recording, data from the entire recording were used for training, validation and testing (i.e., also including sleep episodes lasting longer than 15 s; basically, sleep stage N1) to obtain as much data as possible as the fraction of time covered by MSEs is small (5 to 8 %; Table 1). In total, 1262 MSEs and segments of sleep were scored.

### 2.2    Preprocessing

The signals were bandpass filtered with a Fourier filter in the band 0.5-45 Hz (FFT of EEG followed by setting of frequencies <0.5 Hz and >45 Hz to 0 and then performing an inverse FFT). This step is considered as signal conditioning and is necessary for the application to future data that are recorded with different devices. We still refer to it as raw data as no features were derived for the classification.

For each training sample, we used one occipital EEG derivation and two EOG channels. The EEG derivation for each training sample was chosen randomly out of two derivations (O1M2 or O2M1) and we assigned the corresponding scoring. Thus, we effectively doubled our training set by using both EEG channels as independent signals. Since both EEG signals were similar and most of MSEs were observed in both channels simultaneously we did not gain completely new examples by this procedure, but it served as data augmentation. Data augmentation is commonly referred to slight changes to the data, such as additional noise, cropping or warping. It helps to avoid overfitting of the networks (22). Video recordings were not used for automatic classification. For the validation and testing we detected the events using only EEG channel O1M2, the two EOG channels (E1M1 and E2M1), and the corresponding expert scoring.

## 2.3   ML methods

Many pattern recognition problems are easy to solve for a human expert (for example object recognition in images), but it is incredibly hard to define explicit decision rules for such tasks. Machine learning methods are proven to be very efficient for pattern recognition tasks (13, 23, 24), including EEG data (14-18, 25, 26). The idea behind machine learning is to let the algorithm learn the patterns in the data. This can be achieved either in a supervised way, i.e. when there are labels attached to each datapoint, or an unsupervised way, when there are no labels and the algorithm should find the structure in the data on its own. A typical example of unsupervised learning is clustering (27), and the most common example of supervised learning is classification (23). In this work, we are aiming to detect MSEs. This problem can be solved in different ways. For example, one can solve it as object detection problem (28-32), where the objects are MSEs. Since the MSEs are not overlapping it can also be considered a segmentation problem. Further, we can also represent it as a classification problem for every sample, i.e. we classify each sample of a recording as one of the four classes: wake, MSE, MSEc or ED. We have chosen to use the classification approach.

## 2.4   Classification

We developed and implemented automatic classification algorithms (supervised learning) based on a Convolutional Neural Network (CNN) (33). Such a network uses small filters, and every layer of the network has its own set of filters. Each filter is convolved with an input to the layer, i.e. the filter is moved across the input and a similarity measure is computed for every position and stored in a new matrix. Matrices corresponding to all filters are stacked together and this stack is the input for the next layer. We used small filters based on empirical knowledge. Further, it was shown that deep networks with small filters perform best (34). We used a small number of filters in the first layers of the network and more filters in later layers. It is a common approach used in computer vision (34). Filters in the first layers have a small receptive field and usually detect simple patterns, thus, there is no need for a large number of filters in the first layers. The layers located deeper have larger receptive fields, thus they detect more complex patterns, and it makes sense to increase the number of filters to extract the maximal amount of information from the signal. It is common to use the number of filters equal to a power of two. We used a similar approach and choose the number of filters as in our previous EEG analysis paper (18). Also the same ladder of convolutional layers was used in (34) and other works on computer vision. The number of layers was chosen such that the last layer's receptive field covers the whole input window.

Since we wanted to assign a label to each sample of the signal, we ran the classification algorithm on a sliding window. The stride of the sliding window was equal to the segmentation resolution, i.e. one sample. We could have used a larger stride and predicted a label not for every sample but for example every 100 samples. Resolution would be lower, but computational expenses would be reduced, and the algorithm would be faster. However, we wanted to avoid coarsening of the expert's segmentation resolution (please note that this was done for CNN-LSTM network architecture; see below). Our CNNs predicted the label for the central point of each window. We could minimize the fringe effect in this way, i.e. the different amount of information available at the edge and in the middle of the window. The amount of information available at the edge is lower than in the middle, thus, we chose to work with a sliding window. The idea of using a convolutional neural network on a sliding window is illustrated in Figure 1a and its structure in Figure 2a.

We also implemented a combination of convolutional and recurrent neural networks (RNN; Figure 1b) as RNN take into account the temporal structure of the data (35) to test whether performance could be considerably improved. We wanted the network to see a certain window, it can be achieved either by using a CNN with large input size or a combination of CNN and LSTM. In the latter case we have the same input window size with much smaller number of parameters. We first processed the signals with a CNN with a 1-s window. The windows were overlapping, and the stride was equal to 50 samples (0.25 s). We chose relatively large stride to speed up this network. As a consequence of the large stride we predicted the label every 50 samples and the resulting resolution of the prediction was lower than the resolution of the other networks used. We do not think this is a problem since the MSEs are $1 - 15$ s long by definition. Next, we used a recurrent neural network, namely a long-short term memory (LSTM) (35) network. The LSTM network received the vectors resulting from the CNN as input (Figure 2b) and the output was a sequence of labels (MSE, Wake, ED, or MSEc). Each label was assigned to the center of the corresponding CNN window.

Most of our networks were convolutional networks working with a sliding window (CNN) and one network was a combination of convolutional and LSTM networks (CNN-LSTM network).

## 2.5   Architecture of the networks

Figure 2 (**a** CNN and **b** CNN-LSTM) illustrates the network architectures.

Raw EEG and EOGs (in µV) served as the input data for CNNs and they were divided by 100 and clipped to the range [-1; 1] to keep weights and gradients small. For CNN-LSTM network similar procedure was performed, however, we first added 100 to the signals, divided them by 200 and clipped them in the range [0; 1]. In the first layer of the network we added some Gaussian noise (std = 0.0005) to increase robustness of the network to noise.

Convolutional blocks are the basic parts of the networks. They are composed of convolution followed by batch normalization, activation and max-pooling, i.e. filtering, nonlinear activation and reduction of the size of the tensors (Fig. 2 and Tab. 2).

We first explored different network configurations based on our previous experience with sleep stage scoring (18) and decided to investigate the ones finally implemented in detail. However, the parameter space is infinite, and we do not claim that our choice is the best one. Some of the blocks were repeated many times because we want to make the network deep and would like to end up with a vector of size 1 in the temporal dimension (i.e., the receptive field of the last layer covers the whole input window) and a large size in the dimension of the filters (that these filters can contain large amount of information about the input window). Thus, some of the blocks are repeated different number of times depending on the size of the sliding window, i.e., for each doubling of the window size, we repeated the block one more time to increase the depth of the network accordingly: 3 times for 2-s, 4 times for 4-s, 5 times for 8-s, 6 times for 16-s and 7 times for 32-s windows. In the end we used 5 different window sizes. We limited the length of the sliding window to the range of 2 to 32 s because we explicitly did not want the networks to learn MSE duration criteria, only the underlying EEG patterns. For practical applications, duration criteria can easily be applied post-hoc. We also tested a network (16 s long window) with a single EEG channel as input instead of an EEG channel stacked together with the two EOG channels. To account for the imbalance between the stages (Table 1), weights inversely proportional to the frequency of a class were generally applied. To test for the impact of the weighting, an additional network (16 s long window) was trained with equal weights. This resulted in seven CNN networks and one CNN-LSTM network, in total 8 different network configurations to explore.

The notations used in Figure 2 and the corresponding parameters are summarized in Table 2. For the parameter values applied see the corresponding values in Figure 2.

## 2.6    Performance evaluation

There are several methods to evaluate the performance of a classification algorithm. The simplest one is to find the proportion of correctly classified examples, a metric called accuracy. While it might be a good measure when we have nearly the same number of examples of each class, it is a very poor measure in case the dataset contains predominantly examples of one class. In our case the most frequent class was wakefulness. Imagine that 90% of the data is labeled as wakefulness, then a classifier, which labels all the data as wake would result in 90% accuracy, but such a classifier would be useless.

One can compute measures such as sensitivity and specificity. These measures take into account both true positive and true negative results. In this case we need two numbers to characterize an algorithm. However, it is more convenient to have a single number to measure performance. Many different single-number measures exist but they always capture only partial information about the quality of an algorithm.

We used Cohen's Kappa (36) to measure the quality of the algorithms. This measure compares the output of the classifier with one that would give random answers with the probabilities of classes taken according to the proportion of examples of a corresponding class in the original data.

The main disadvantage of Cohen's Kappa is the fact that if our data contains only one class, kappa will be equal to zero. For example, a kappa for a particular subject who was always awake, and the algorithm correctly classified the entire recording as wake will be equal to zero. This would indicate a very bad performance, despite the fact, that such a segmentation is correct.

There are two important aspects regarding the computation of Cohen's Kappa in this work. First, we could not compute kappa for each patient since in some recordings not all classes were present. Thus, we concatenated all the recordings and then computed kappa resulting in an overall performance. As a consequence, error bars are not available. Second, we computed kappa for each class separately. In order to compute kappa for a particular class k, we assigned the labels of the examples of the class k to 1 and all other labels to 0 and then computed kappa. We repeated this step for each class.

## 2.7    Training, validation and testing

As mentioned above, our data comprised 76 MWT recordings, one recording per patient. The data was split into three parts: 70% training (n=53), 15% validation (n=12) and 15% testing (n=11). Only the best performing network was additionally evaluated using the test dataset. The demographic data, diagnosis and fraction of time spent in the four stages of the patients contributing to the three parts are provided in Table 1. Most of the time the patients were awake (85 to 91 % of the time) and in 5 to 9 % of the time MSEs occurred.

We used the Keras package (v 2.2.0) (37) with the Tensorflow (v 1.8.0) (38) backend to train the networks and Python 3.5.2 to run the scripts. Data conversion and filtering was performed with Matlab 2018b.

We trained the networks using the Adam (Adaptive momentum estimation) optimization algorithm (39) with Nesterov momentum (40) (Nadam in Keras with the default parameters, learning rate 0.002). For the CNN-LSTM network gradient clipping at a value of the gradient norm equal to 1 was applied.

The batch size (stack of input windows) for CNN networks was equal to 200 and 128 for CNN-LSTM network. The input windows were selected randomly for each batch without repetitions.

We trained every CNN network for 3 training iterations and the CNN-LSTM network for 8 iterations. Here we use the term training iteration instead of commonly used training epoch because epoch is reserved for scoring epoch in the literature on sleep analyses. It appeared that the performance reached its maximum already after only one training iteration and did not improve further. This is not surprising given that our dataset included a frame for every sample of the signal. It produced a lot of redundant data because the frames corresponding to consecutive samples differ only in the first and the last values and thus are almost identical. Thus, our networks were able to converge within one training iteration.

## 2.8    Visualization

Our data contained 4 classes defined by an expert and it was interesting to see how they are represented in the feature space. We took the best performing network (with 3 input channels and a 16-s window) as we used it for solving the classification problem and added one more convolutional layer with 64 filters of size 3. The reason to use an additional layer was to reduce the size of the resulting feature vector. We used the output of the last convolutional layer as a feature vector. The length of the vector was 64, which is large. Thus, it was not realistic to look at the data points in this 64-dimensional space. Fortunately, there are many dimensionality reduction methods available. We have chosen the t-distributed stochastic neighbor embedding (t-SNE) (21) to project the data into a 2D space. This mapping preserves the distance ratios between the data points. In this way we can see whether separable clusters of data points exist. It should, however, be kept in mind that this mapping is reflecting the representation of the data by the network (internal representation) and not any sort of ground truth. Thus, the visualization might differ if another network structure is employed.

# 3    RESULTS

## 3.1    How our algorithms performed in classification

Detection of the different classes in one recording with one of the networks (CNN 16s) and the corresponding expert scoring are illustrated in Figure 3. A good match between the algorithm and the expert scoring for wakefulness and MSEs can be seen, but the detection of MSEc and ED was not successful. Performance of the network on the other patients in the validation dataset are illustrated in Supplementary Figure S1 and of patients of the test set in Supplementary Figure S2.

Our algorithms resulted in Cohen's Kappa coefficients close to the ones resulting from the scoring of two experts (5 recordings were scored by two experts; Figure 4). Importantly, our algorithms did not produce any substantial amount of false positive MSE detections in most of the recordings (except one recording). A small amount of false positives (high precision) is especially important for recordings, which do not contain any MSEs.

Cohen's Kappa of the algorithms and of the interrater agreement was good for MSEs and wakefulness (~0.7), but negligibly low for MSEc and ED (<0.1). The results for the different network configurations are illustrated in Figure 4 and summarized in Table 3. We suggest that the CNN with a 16-s window is an optimal network, as we did not observe any further improvement with a 32-s window (Figure 4 and Table 3).

The agreement between the experts for MSEc was higher than the agreement between the algorithm (CNN 16s) and an expert (MSEc - 0.04). Kappa for ED was the same (0.06) when computed between experts and between the algorithm and an expert. Cohen's kappa for both MSEc and ED was very low (<0.1) for both interrater comparison and the comparison of an algorithm with an expert. Such level of agreement is negligible (41). There were five recordings in the validation dataset which contained a very small amount of MSEs or none at all (Supplementary Figure S1). The CNN with a 16-s window detected a substantial amount of false positive MSEs in one of the patients (recording uXdB).

The performance of MSE detection with the best of our CNNs was slightly better than the one with the CNN-LSTM architecture. It might be due to different resolution of detection. We cannot be sure that this result would hold if the temporal resolution had been the same. The quality of segmentation was dependent on the length of the window. We think that the optimal length of the window is 16 s, since we did not see further improvements with a 32 s long window. The network with uniformly weighted classes (CNN 16s_u; Figure 4) did not perform better than the ones with balanced weights. The CNN which did not use the ocular channels as an input, i.e. used only a single EEG channel as input, performed worse than a similar network with three input channels (1 EEG and 2 EOG). This suggests that ocular channels contain information important for the MSE detection, most likely slow eye movements, eye blinks and saccades.

We evaluated only the best (optimal) performing algorithm, the CNN with a 16-s window with the test dataset. Evaluation resulted in the following Cohen's kappa values: W - 0.59; MSE - 0.69; MSEc - 0.05; ED - 0.11. These results were very close to the ones resulting from the validation dataset (Table 3), and thus suggest, that there was no substantial overfitting to the validation dataset. Again, we observed no substantial false positive MSE detections in the test dataset, except for one recording (patient f8H5; Supplementary Figure S2). Overall there were six recordings with no or very little MSEs in the test dataset and five of them were scored nearly perfectly by the algorithm (CNN 16s). Moreover, the recordings with a substantial amount of MSEs were scored with very high quality (Supplementary Figure S2).

## 3.2   Why did the algorithm not perform equally well for all classes?

Visualization (t-SNE) and analysis of the internal representation of the data in our network (CNN 16s) revealed as expected for the training of artificial neural networks, that in the representation of training data all four stages form clearly separated clusters except for very few data points (Figure 5a; Supplementary Figure S3). However, in the representation of validation data generally the four classes are not separable. In most cases there were two clear clusters representing wakefulness and MSEs with a smooth transition between them (Figure 5b; Supplementary Figure S4). However, most MSEc and ED were on the interface between these two classes, which explains why they cannot be reliably identified by the algorithm. In some cases (Supplementary Figure S3; patient IhpU), we observed not only a cloud of MSEs, which was connected with the cloud of wakefulness but additionally a second clearly separable cluster. This distinct cluster may not represent MSEs but sleep episodes longer than 15 s which were marked as microsleep by the expert (see Discussion).

## 4   DISCUSSION

Our algorithms reliably identified MSEs and wakefulness with a performance close to a human expert and did not produce any substantial amount of false positive MSEs detection in recordings of patients, indicating that reliable automatic MSE detection is feasible based on raw EEG and EOG data recorded during the MWT in a clinical setting. In one of the recordings (uXdB; Supplementary Figure S2) we observed a considerable amount of false positive MSE detections. We do not yet have an explanation why this happened. Visual inspection of recording uXdB revealed that it was quite noisy. Thus, it would make sense to test the algorithm on more data, especially noisy ones to check if noise poses a problem for the algorithm.

We provide a proof of principle that reliable automatic detection of MSEs using raw data is feasible and of a high quality. The performance of the CNN with a 16-s window on validation and test data was very similar indicating that there was no substantial overfitting, and the algorithm performs well independent of any disorder or medication. However, we would need more recordings double scored by independent experts, and overall larger datasets to draw a final conclusion. Further, evaluation of the algorithms on data of healthy subjects and subjects recorded in a driving simulator should be performed (42).

Performance of our raw data based approach was similar to the feature-based ones (11). The feature-based algorithms of Skorucak et al. (11) detected only bilateral occurring MSEs, i.e. MSEs occurring in both occipital EEG channels simultaneously and was not trained to detect MSEc and ED. It is easier to detect MSEs occurring bilaterally (two channels) than detecting them based on a single channel. Moreover, the feature-based algorithms worked with a 0.2-s resolution, and the subdivision into training and testing data sets were different, i.e., randomization of individuals was different and there was only a test set (no subdivision into validation and test). Thus, a direct comparison of the algorithms must be made with care. The feature based artificial neural network (11) detected wake and MSEs with kappa values of 0.65 and 0.75, respectively (calculated as described in 2.6 Performance evaluation). The same algorithm applied to MWT recordings of healthy subjects after sleep deprivation (42) revealed kappa values of 0.61 and 0.65, respectively. Taken together, with our approach (best performing 16-s network) we achieved a similar performance (validation: 0.67 and 0.69; testing: 0.59 and 0.69). In the feature-based approach (11), EEG recordings had first to be cleaned of electrocardiography (ECG) artifacts to be able to reliably classify the data as the features were mainly derived from EEG spectra. Human scorers, however, were not distracted by these artifacts. Similarly, our raw data based approached worked well without prior ECG artifact removal. Generally, we expect that raw data based algorithms would be more robust and better transferable to other datasets and might be better suited for on-line processing.

Performance of the CNN algorithms depended on the length of the sliding window. We think that 16 s is an optimal window size because we did not observe further improvement with a 32-s compared to a 16-s window. Even the network with a 2 s long window performed reasonably well. This is an interesting observation because an expert needs to see 10 to 20 s of the signals to score MSEs. Further, training with a 16-s window and weights inversely proportional to the stage prevalence or with equal weights resulted in the same performance (Fig. 4 and Table 3) indicating that the low prevalence of MSEs (Table 1) was not an issue; MSEc and ED could not be detected with bot weightings.

One EEG and two EOG channels served as input of the classifiers, except for one case. Classification based on a single EEG derivation (16s_1c) worked well, suggesting that the occipital EEG contains substantial information to score MSEs at least for our conservatively defined MSEs as short as 1 s (7). Nevertheless, a similar network, which used also EOG signals as input, performed better. This was expected since the eye closure is a criterion for expert scoring. Moreover, eye blinks or saccades might be correlated with wakefulness providing additional information for the algorithm.

Borderline segments between clear wakefulness and MSEs that were particularly difficult to score were categorized as MSEc or as ED (7) in the BERN microsleep scoring criteria. Both, experts and algorithms performed bad in scoring these borderline segments (Figure 4; Table 3). After visualizing the internal representation of the data in the neural network we came up with a hypothesis why it might be the case (Figure 5 and Supplementary Figures S3 and S4). Visualization (t-SNE) of the internal representation of the data in one of the networks (CNN 16s) revealed that generally the 4 classes were not completely separable. In most cases there was a smooth transition between the clusters of wakefulness and of MSEs (Figure 5b, and supporting information Figure S4). Most MSEc and ED were at the interface between MSE and wake and overlapped with them considerably. This explains why they cannot be reliably identified neither by the algorithm nor by an expert. Thus, in contrast to MSEs, MSEc and ED are currently far from being practically applicable. Please note that this visualization only reflects representation of the data in the particular neural network. For other networks the representation might be different.

In some cases (Supplementary Figure S3; patient IhpU), we observed not only a cluster of MSEs, which was connected with the cluster of wakefulness but also a second clearly separable cluster of MSEs. These distinct clusters may not represent MSEs, but sleep episodes longer than 15 s (stage 1), which were marked as MSEs by the expert as the occurrence of consolidated sleep was missed by the technician and the recording continued leading to MSEs lasting longer than 15 s. Note, that we observed such a cluster only in the training dataset. We did not observe this in the validation dataset but observed several clusters of points marked as wakefulness Supplementary Figure S4).

Cohen's kappa was somewhat higher for the inter-rater agreement. However, it is important to note that the interrater agreement was assessed on only five recordings, which were not selected completely randomly. The experts randomly selected only recordings, which contained MSEs. Moreover, the experts were trained in the same laboratory and the second expert checked the scoring of the first one for about 2/3 of the recordings.

Our CNNs performed classification for every sample, thus the detected episodes are likely to be fragmented. This issue can be easily solved with median filtering or splitting the results into consecutive intervals and assigning the most frequent class to all samples in the corresponding interval. We used latter approach for the visualization in Figure 3 using 0.5 s long intervals. Additionally, classification was performed based on a sliding window shifted by one sample.

The use of occipital EEG channels was based on clinical experience since features of the MSEs are often best visible in this brain region (7). In particular, the alpha rhythm observed during rest with eyes closed is best observed over occipital brain areas. Further, the transition to sleep is accompanied by a slowing of the EEG, i.e., a loss of alpha activity and a shift to theta activity which again is best seen in occipital derivations (7). Given the local aspects of sleep, future development of algorithms should take other brain regions into account.

As a result of this work, we provide a proof of principle that reliable automatic MSE detection with deep neuronal networks working with raw EEG and EOG data as input is feasible with a quality close to the one of human experts. Deep neural networks may also be used as a tool to visualize data and thus, foster their interpretation and gain new insights.

## 5    CONFLICT OF INTEREST

Authors declare no conflicts of interests.

## 6    AUTHOR CONTRIBUTIONS

AM developed and programmed the algorithms and conducted the analysis. AM and PA wrote the paper. AHG, DRS, and JM brought in longstanding expertise with microsleep episodes in patients, collected the data and performed visual scoring. JS provided a script for reading the scoring information and her expertise with the data. PA, JM, and DRS organized funding. All authors commented on the manuscript, proposed corrections and agreed on the final version of the paper.

# 7    FUNDING

This work was supported by the Swiss National Science Foundation (grant 32003B_176323), nano-tera.ch (grant 20NA21_145929), the Clinical Research Priority Program "Sleep and Health" of the University of Zurich, and the Swiss Commission of Technology and Innovation (CTI; grant 17864.1 PFLS-LS). NVIDIA Corporation provided us a GPU in the framework of academic grants.

# 8    ETHICS

The study was conducted according to the Declaration of Helsinki, Swiss Law, and ethical approval of the local ethics committee (KEK-Nr. 308/15). Data were included based on a general consent which the patients signed with the hospital.

# 9    DATA AND CODE AVAILABILITY

The code of our algorithms is available at https://github.com/alexander-malafeev/microsleep-detection .

MWT data and the corresponding scoring are available in the Zenodo repository https://zenodo.org/record/3251716 (DOI: 10.5281/zenodo.3251716).

## 11    FIGURES

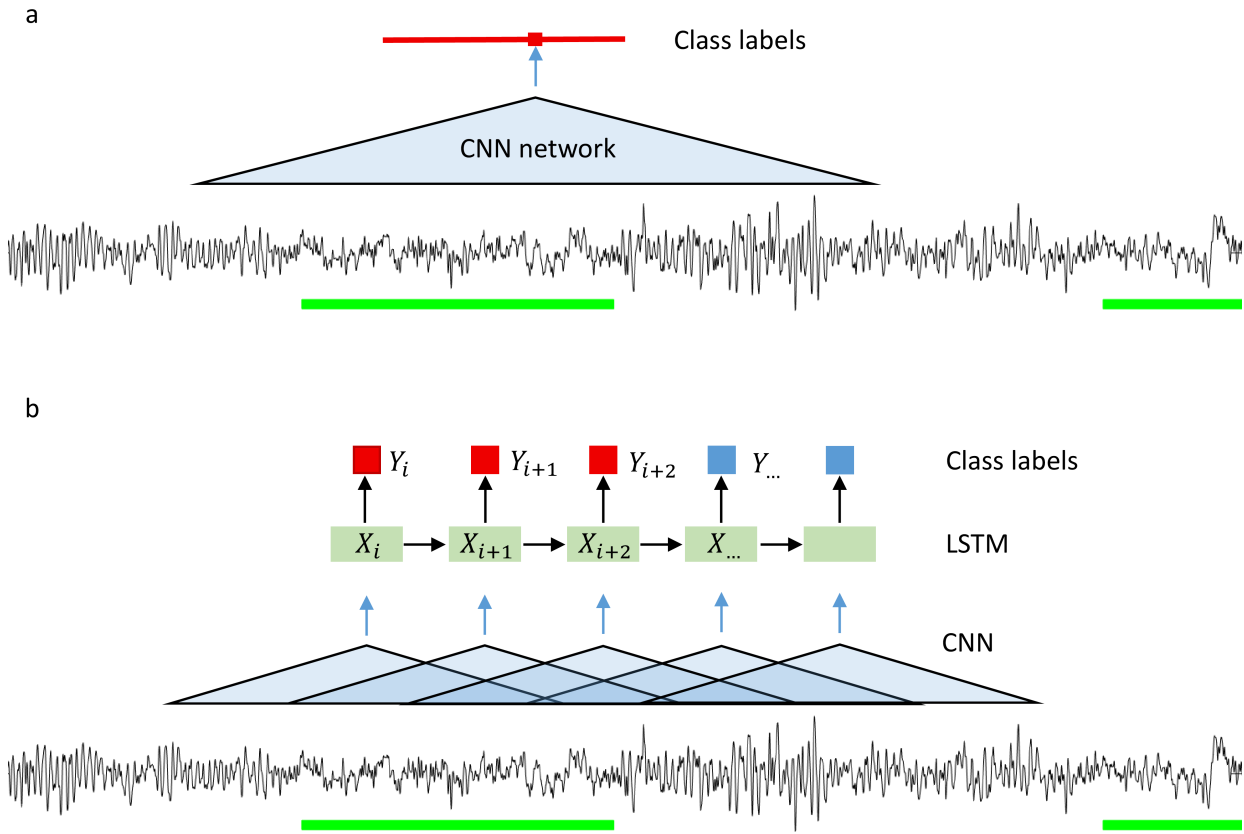

**Figure 1**. Illustration of the idea behind the segmentation (classification) with a CNN (a) and CNN-LSTM (b) network. A sliding window was used in the case of CNN only networks. Microsleep episodes were inferred corresponding to the middle of the window on every step (sampling resolution). In case of CNN-LSTM network a sequence of overlapping windows (classified by a CNN) with the stride of 0.25 s inferring microsleep corresponding to the middle window in the sequence (LSTM classification). Thus, the resulting resolution of the detection was 0.25 s. Green bars: scored MSEs; red bar in (a): classified MSE; red and blue squares in (b): classification of a sample or window.

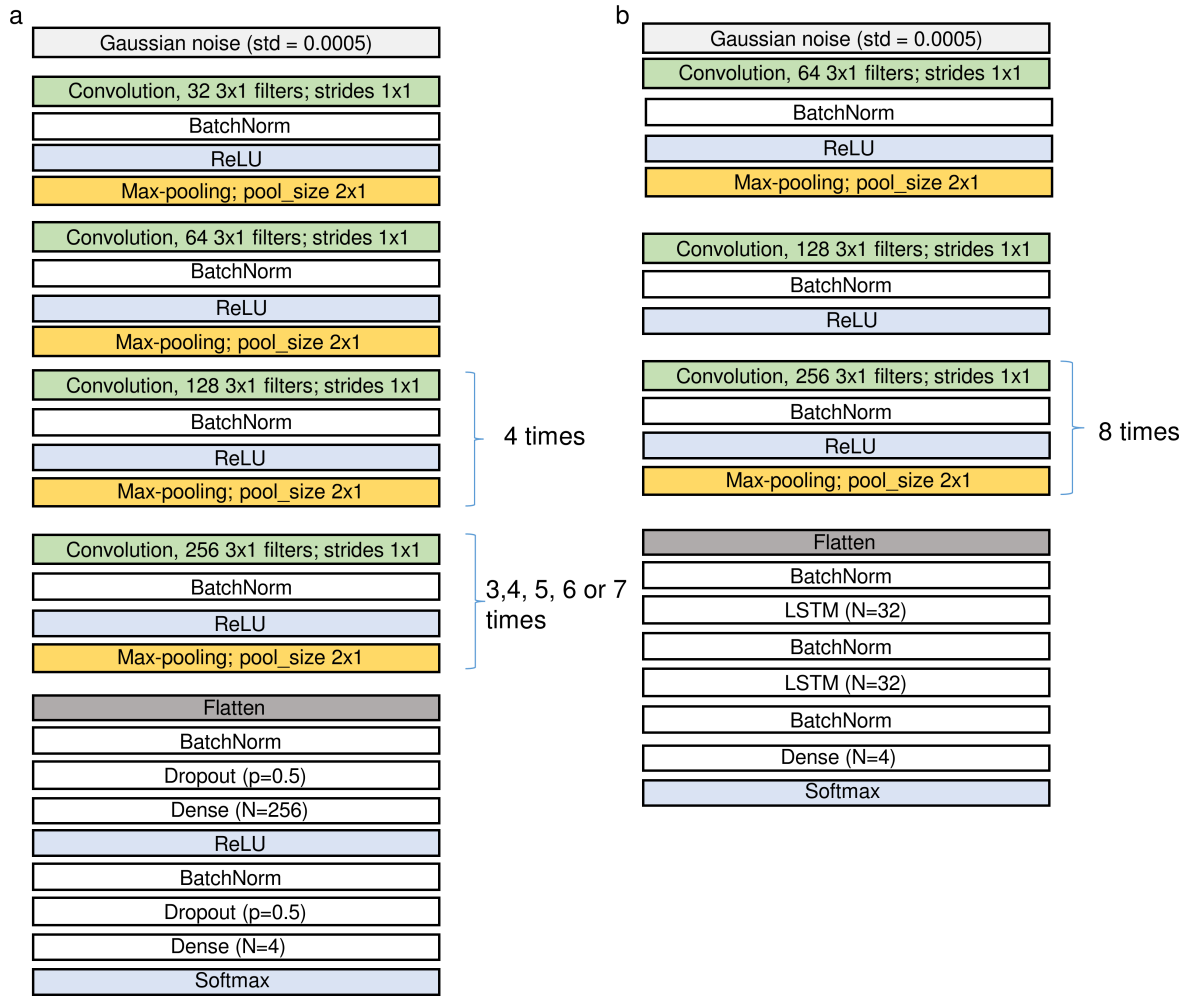

**Figure 2**. Structure of the CNN (a) and the CNN-LSTM (b) networks. Input is on top, output at the bottom. Since we applied several configurations of the CNN networks the repetitions of the last convolutional and pooling blocks were different. The number of channels in the input may differ for the networks using either an EEG and two EOG channels or a single EEG channel only. See Architecture of the networks in Methods and Table 2 for the description of the different layers.

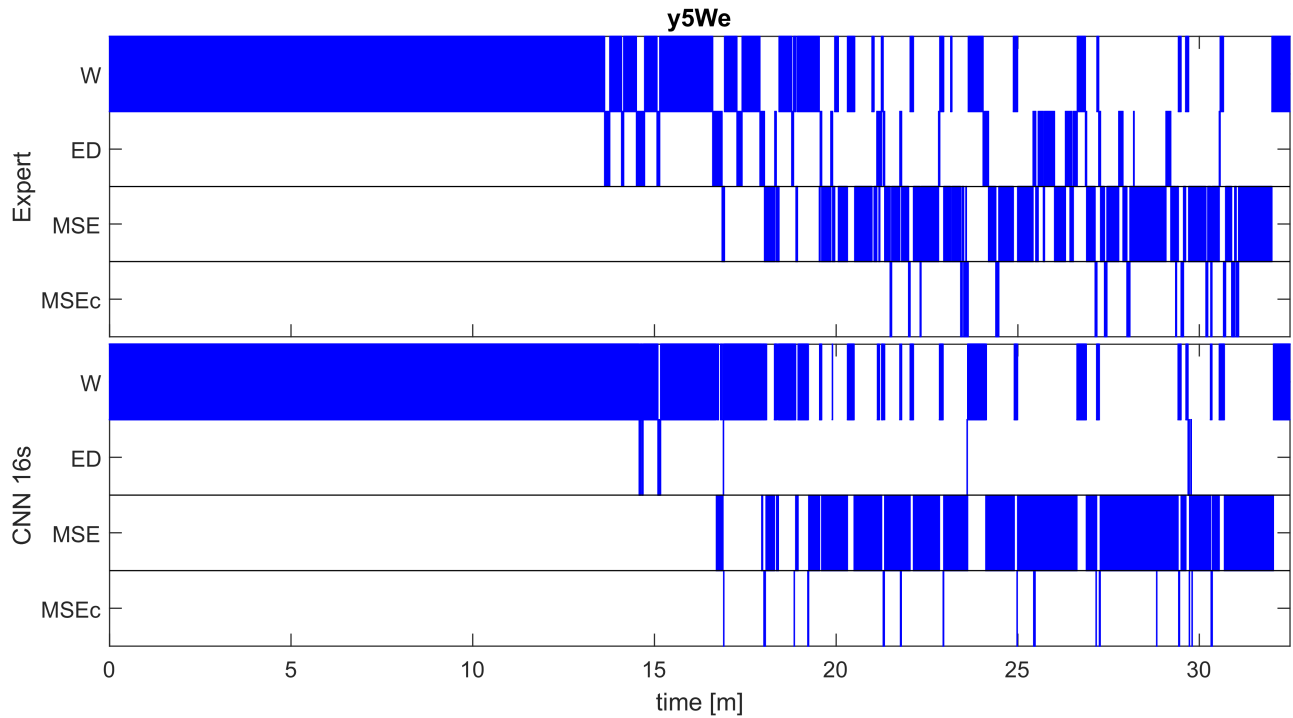

**Figure 3**. Expert (top) and automatic soring with one algorithm (CNN with 16-s window; bottom) of an MWT (40 min) in one patient of the validation set (patient y5We). A good match between the algorithm and an expert scoring for wakefulness (W) and microsleep episodes (MSE) are evident, but a poor match for episodes of drowsiness (ED) and microsleep episode candidates (MSEc). Scoring was performed with the resolution of one sample; for the illustration, we coarsened the result to a resolution of 0.5 s (100 samples), i.e. the most frequent class within an interval was plotted. Results for other patients of the validation set are illustrated in Supplementary Figure S1, those of the test set in Supplementary Figure S2.

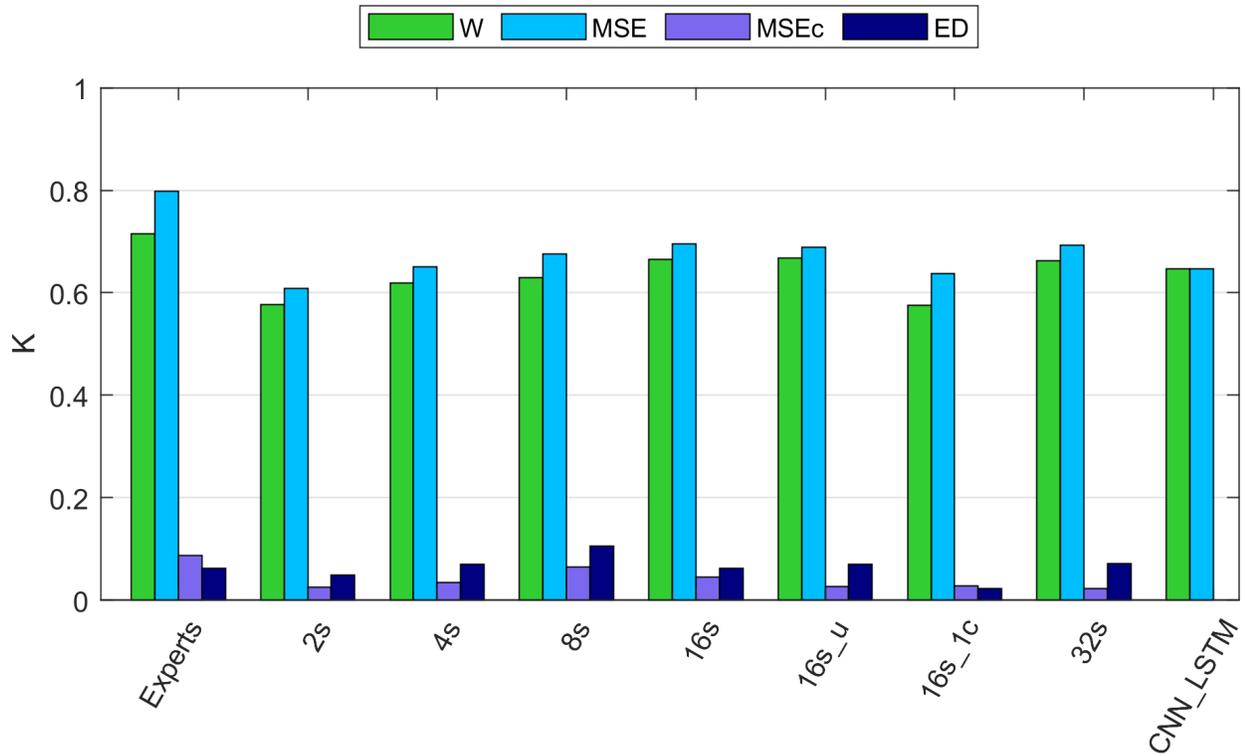

**Figure 4.** Cohen's kappa of different algorithms along with the agreement between two experts. W: wakefulness; MSE: microsleep episodes; MSEc: microsleep candidates; ED: episodes of drowsiness. Experts: agreement between two experts computed based on five recordings containing MSEs. 2s to 16s: comparison between one expert and convolutional neural networks (CNNs) with window lengths 2, 4, 8 and 16 s. 16s_u: CNN with a 16-s window and uniformly weighted classes. 16s_1c: CNN with 16-s window and only one EEG channel as input. 32s: CNN with a 32-s window. CNN_LSTM: CNN combined with a long-short term memory (LSTM) architecture; it has only two classes because this network was trained to detect only MSEs, everything else was considered as wakefulness. If not mentioned otherwise, one occipital EEG channel and two ocular channels served as input for the networks. Kappa of the neural networks was computed using the validation dataset (12 recordings). The data of all recordings were concatenated to estimate the overall kappa.

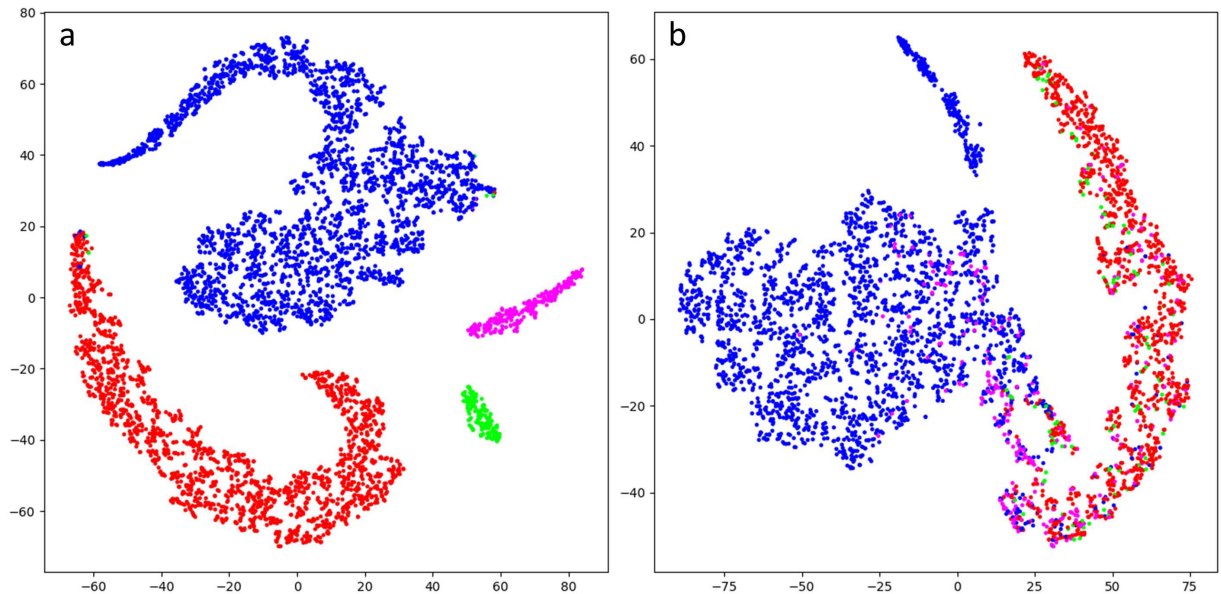

**Figure 5**. T-distributed stochastic neighbor embedding (t-SNE) was used to illustrate the data and their classification mapped into a 2D space (last layer of the CNN 16s; arbitrary units). *(a)* Mapping of training data (patient Nzhl). All stages form clearly separated clusters except for very few data points as expected for the training of artificial neural networks. *(b)* Mapping of validation data (patient y5We; same data as in Figure 3). Basically, two large clusters corresponding to W and MSE are visible which do not completely separate. MSEc and ED do not form clusters and are not separable from W and MSE. Thus, it illustrates why our algorithms could not score MSEc and ED reliably. Wakefulness (W): blue; microsleep episodes (MSE): red; microsleep episode candidates (MSEc) green; episodes of drowsiness (ED): magenta. For the convenience we illustrated only every hundredth datapoint (sample). Please note that this figure only shows the internal representation of the data in this specific network. Further data are illustrated in supporting information, for training and validation separately (Supplementary Figures S3 and S4).

## 12    TABLES

|  | Training | Validation | Testing |
|---|---|---|---|
| N | 53 | 12 | 11 |
| Male/Female | 35/18 | 6/6 | 9/2 |
| Age (mean ± SD years) | 45.99 ± 18.17 | 44.64 ± 20.56 | 44.92 ± 14.48 |
| Sleep apnea | 20 | 0 | 3 |
| Idiopathic hypersomnia | 2 | 1 | 1 |
| Non-organic hypersomnia | 1 | 0 | 0 |
| Narcolepsy | 4 | 1 | 1 |
| Insomnia | 1 | 0 | 0 |
| EDS with unclear cause | 18 | 4 | 4 |
| Excessive tiredness | 2 | 3 | 2 |
| Others | 5 | 3 | 0 |
| Fraction of time in |  |  |  |
| wake | 0.89 | 0.85 | 0.91 |
| MSEs | 0.08 | 0.09 | 0.05 |
| MSEc | 0.01 | 0.01 | 0.01 |
| ED | 0.02 | 0.05 | 0.03 |

**Table 1:** Demographics, diagnosis and fraction of time spent in the four stages of patients contributing to the training, validation, and test set: total number of patients (N), number of males/females, mean age and standard deviation, number of patients with a suspected diagnosis of sleep apnea, idiopathic hypersomnia, nonorganic hypersomnia, narcolepsy, insomnia, EDS with unclear cause, excessive tiredness, and others, and the fraction of time spent in wake, MSEs, MSEc, and ED.

| Layers | Description |
|--------|-------------|
| Convolution, N 3x1 filters; strides 1x1 | Convolutional layer (33) with N filters of size 3x1, i.e. one-dimensional filters of the length 3 and the convolution had a stride of length 1. The weights of convolutional filters were initialized with a Glorot normal distribution (43). |
| BatchNorm | Batch normalization is a way to speed up training and regularize the network (44). |
| ReLU | Rectified linear unit (45), a non-linear activation function. It makes the activations of a network sparse and prevents vanishing of the gradients (45). |
| Max-pooling; pool_size 2x1 | Max-pooling layer (46) with pooling size 2. It takes a maximum out of every 2 elements of a tensor. Thus, the size of the resulting tensor will be reduced by a factor of 2. Max-pooling allows us to reduce the size of the vector, retain most useful information and it also has the property of shift invariance. |
| Flatten | Layer which resizes the input tensor and produces a one-dimensional vector with the same number of elements. |
| Dropout (p = q) | Dropout layer (47). It switches off a fraction q of the neurons in the previous layer in the training phase. Dropout is a good way to regularize the networks, i.e. prevent overfitting (47). |
| Dense (N = n) | Densely connected layer with n neurons. |
| Softmax (N = n) | Densely connected layer with n neurons and a special activation function which produces a probability distribution with n values (23). The sum of these values is equal to 1, n is equal to number of classes we want to predict (in our case it was 4) and every output value is the probability that the sample belongs to the corresponding class. |
| LSTM (N = n) | Long short-term memory layer (35) with the size of hidden states equal to n. It has memory and can use information about the past to make decisions in a current timepoint. |

**Table 2.** Description of the different layers and notions used in the architecture of the networks (Figure 2). For the parameters applied see the corresponding values in Figure 2.

|          | W    | MSE  | MSEc | ED   |
|----------|------|------|------|------|
| Experts  | 0.71 | 0.80 | 0.09 | 0.06 |
| 2s       | 0.58 | 0.61 | 0.02 | 0.05 |
| 4s       | 0.62 | 0.65 | 0.03 | 0.07 |
| 8s       | 0.63 | 0.67 | 0.07 | 0.11 |
| 16s      | 0.67 | 0.69 | 0.04 | 0.06 |
| 16s_u    | 0.67 | 0.69 | 0.03 | 0.07 |
| 16s_1c   | 0.58 | 0.64 | 0.03 | 0.02 |
| 32s      | 0.66 | 0.69 | 0.02 | 0.07 |
| CNN_LSTM | 0.65 | 0.65 |      |      |

**Table 3.** Cohen's kappa computed on the validation dataset (n=12) using different network architectures. See Figure 4 for the meaning of the network labels

# 1    Supplementary Figures

**Supplementary Figure S1.** Expert (top) and automatic soring with one algorithm (CNN with 16-s window; bottom) of the 12 patients in the validation set. Scoring was performed with the resolution of one sample; for the illustration, we coarsened the result to a resolution of 0.5 s (100 samples), i.e. the most frequent class within an interval was plotted. The MWT lasted 40 min and was supposed to be terminated earlier if three consecutive 30-s epochs of N1 or one epoch of any other sleep stage occurred. The time axis is in minutes till the termination of the MWT. W: wakefulness; MSE: microsleep episodes; ED: episodes of drowsiness; MSEc: microsleep episode candidates. The patient ID is provided at the top of the plot. Patients of the test set are illustrated in Supplementary Figure S2.

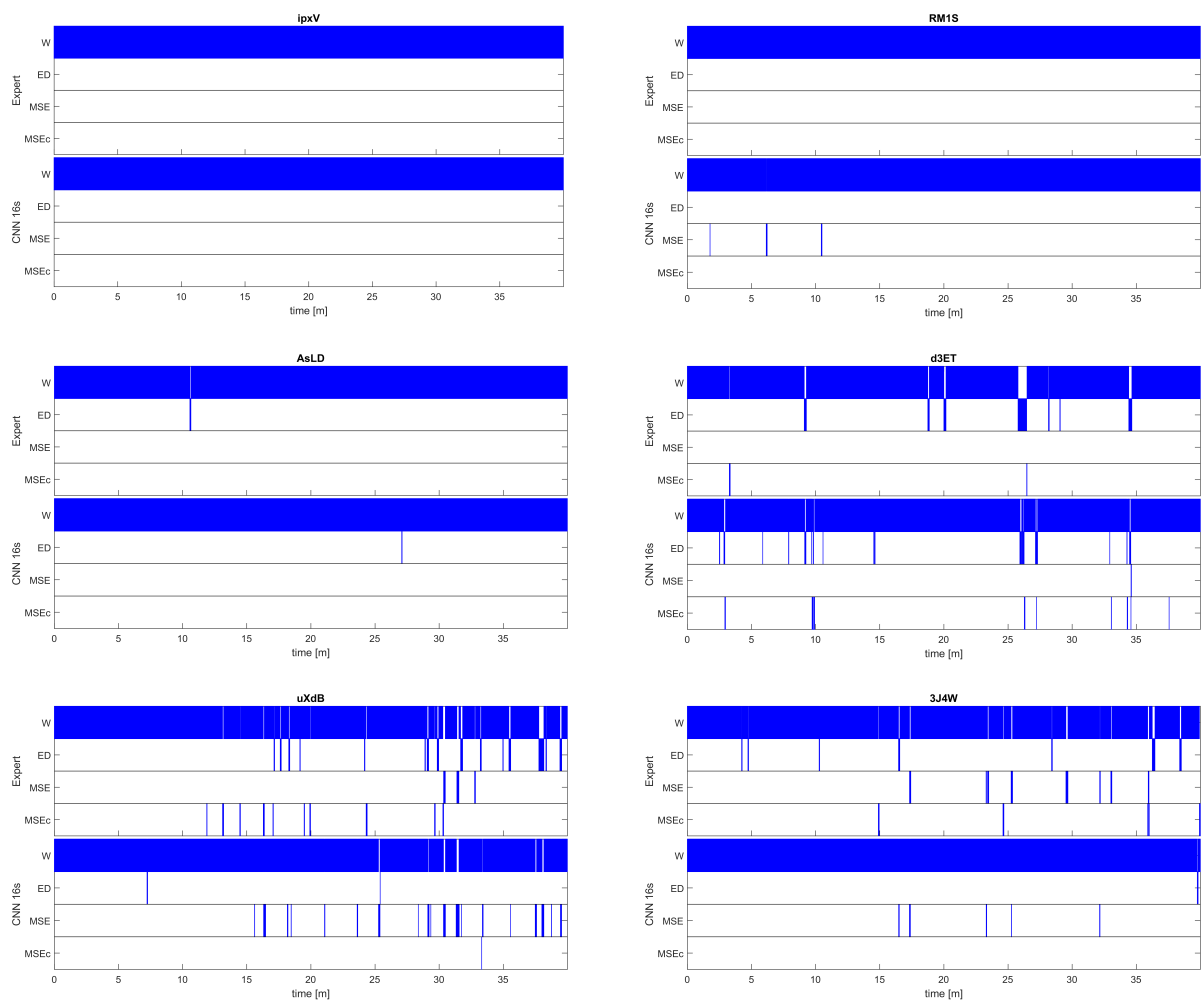

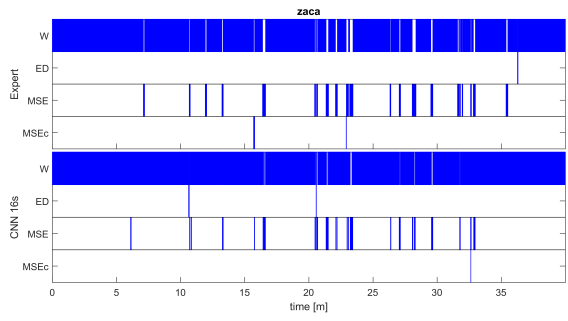

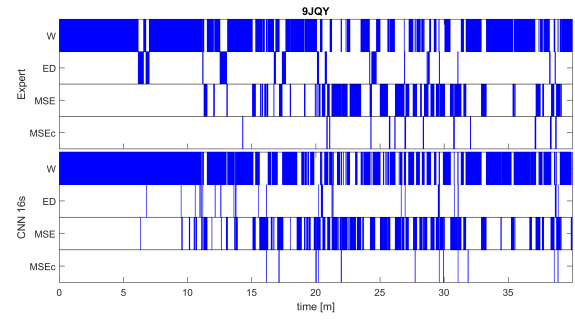

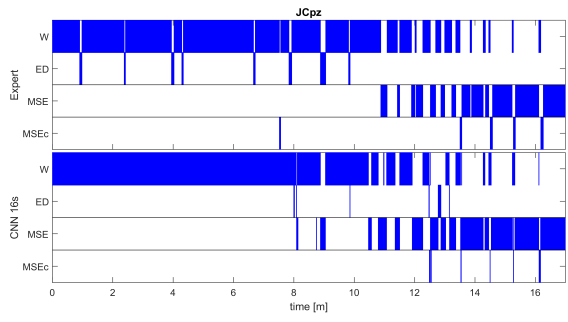

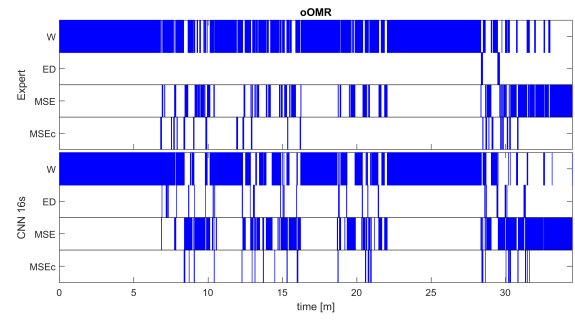

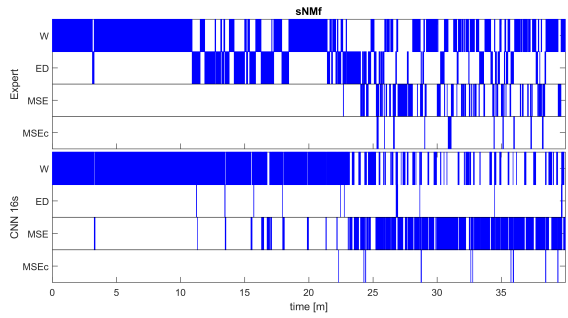

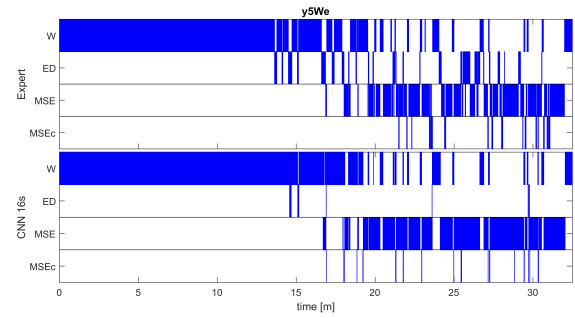

**Supplementary Figure S2.** Expert (top) and automatic soring with one algorithm (CNN with 16-s window; bottom) of the 11 patients in the test set. For details see Supplementary Figure S1.

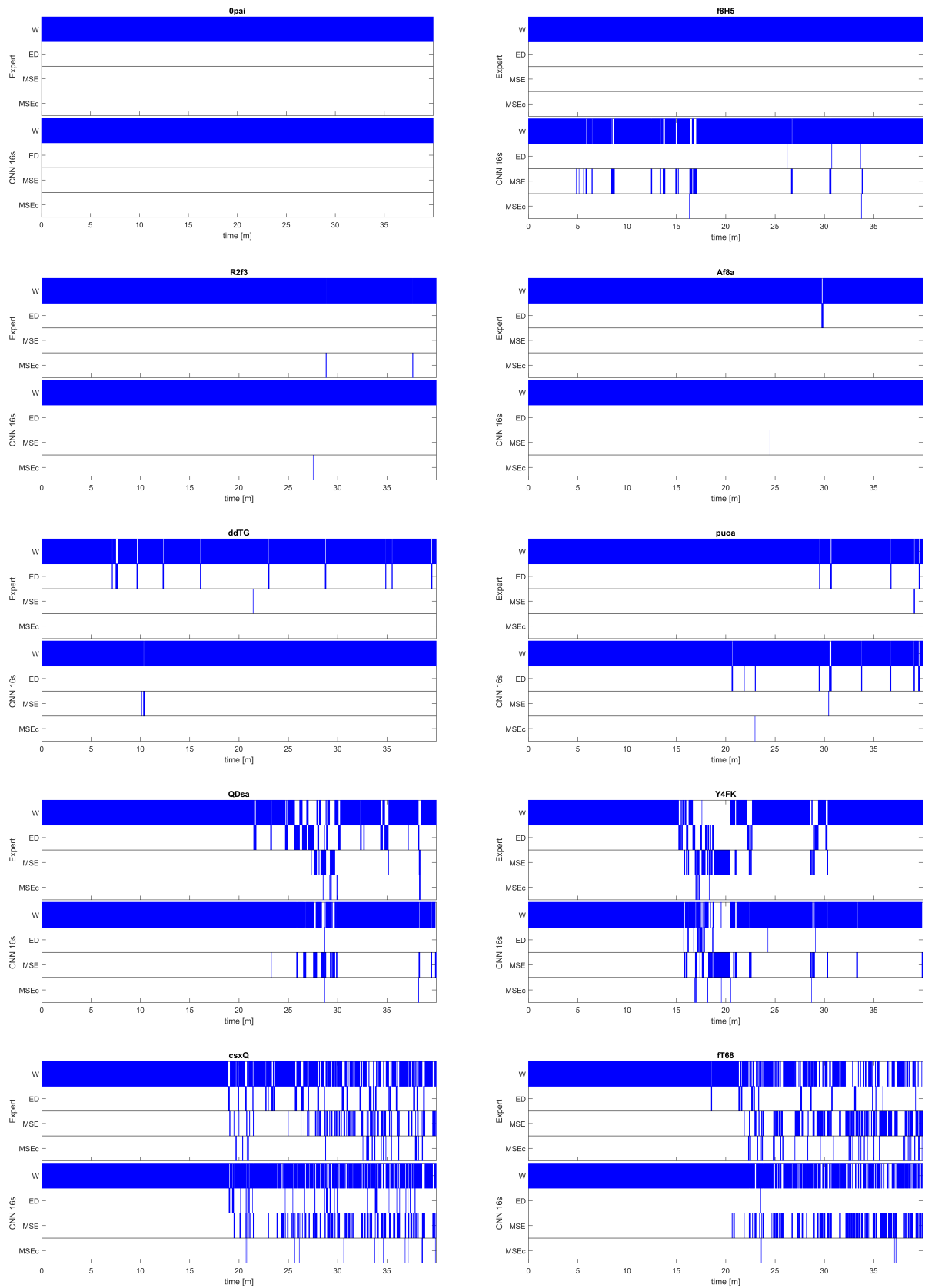

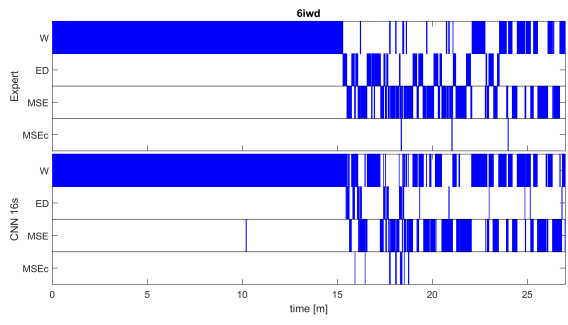

**Supplementary Figure S3.** T-distributed stochastic neighbor embedding (t-SNE) was used to illustrate the 53 patients of the training dataset and their classification mapped into a 2D space (last layer of the CNN 16s; arbitrary units). As to be expected for training data, all stages form clearly separated clusters except for very few data points. Wakefulness (W): blue; microsleep episodes (MSE): red; microsleep episode candidates (MSEc) green; episodes of drowsiness (ED): magenta. For the convenience we illustrated only every hundredth datapoint (sample). The patient ID is provided at the top of the plot. Please note that these figures only show the internal representation of the data in our specific network. Validation data are illustrated in Supplementary Figure S4.

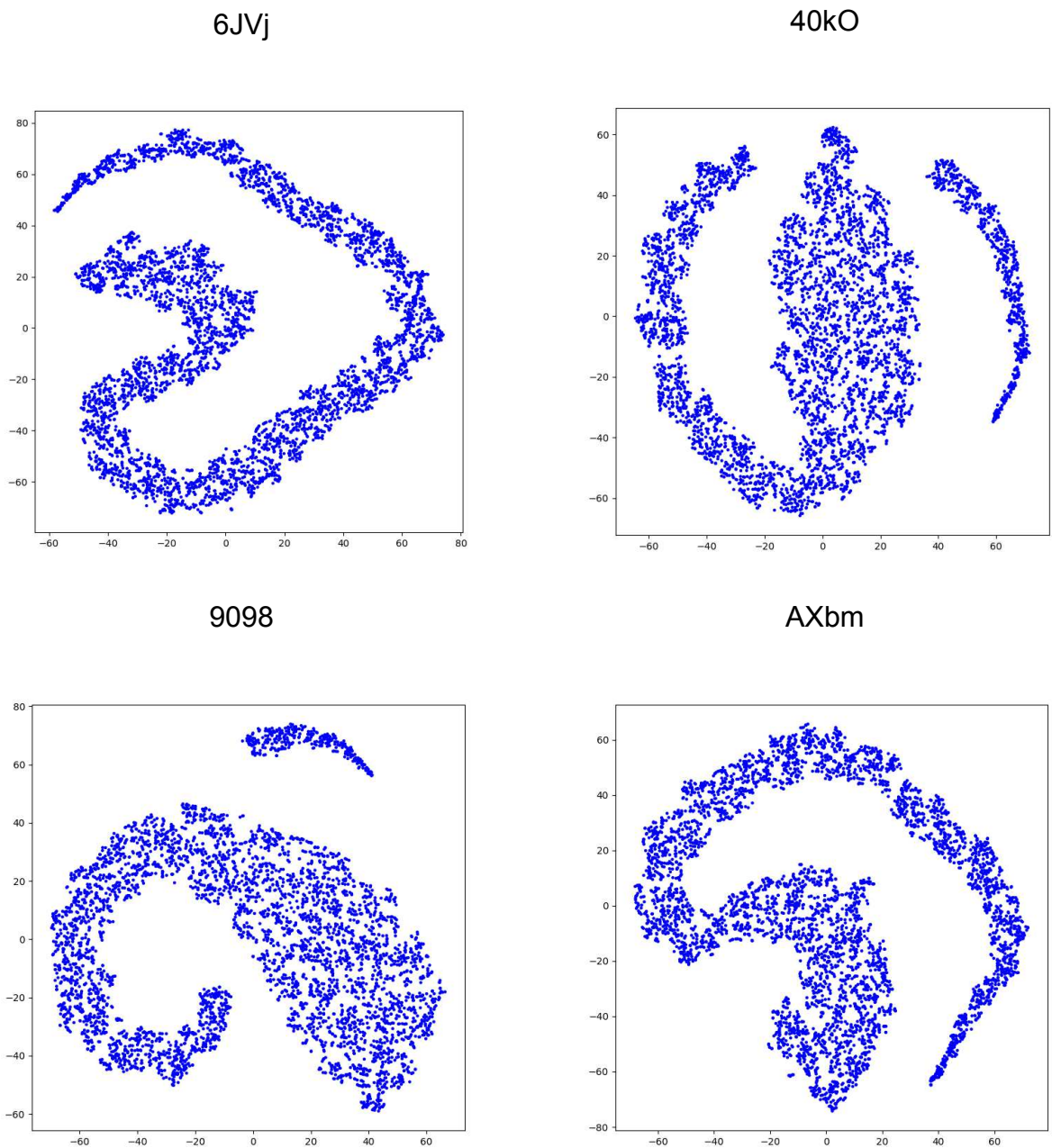

cblr

DSfb

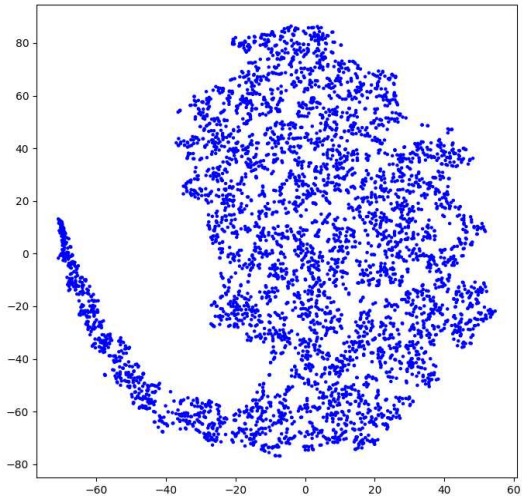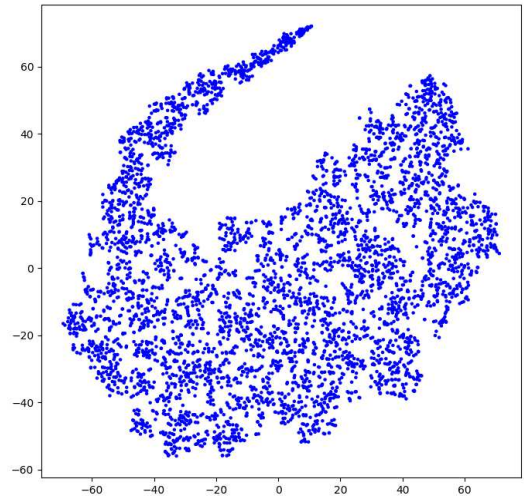

EyTS

mBks

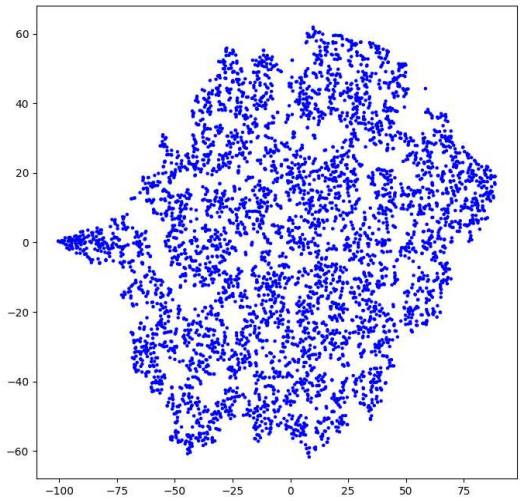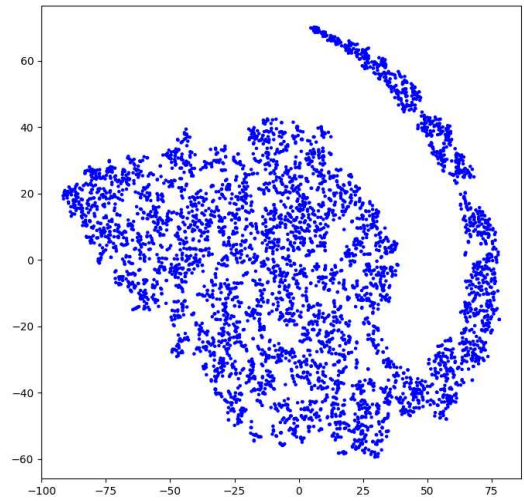

fNe4

go56

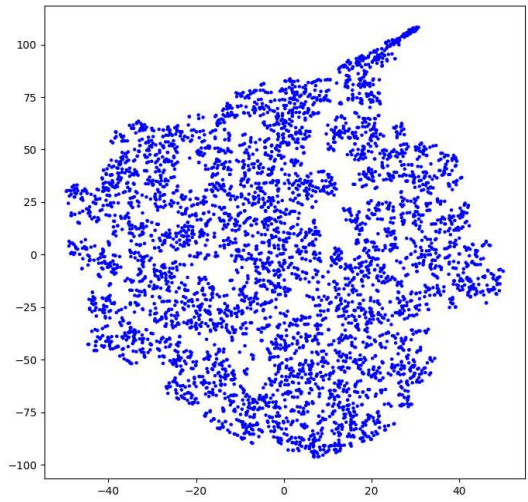

hcml

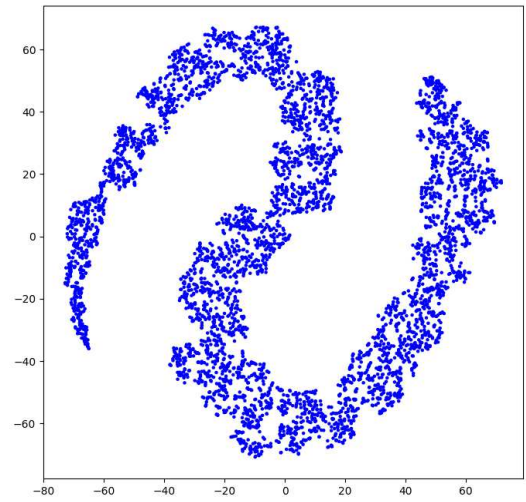

hRMy

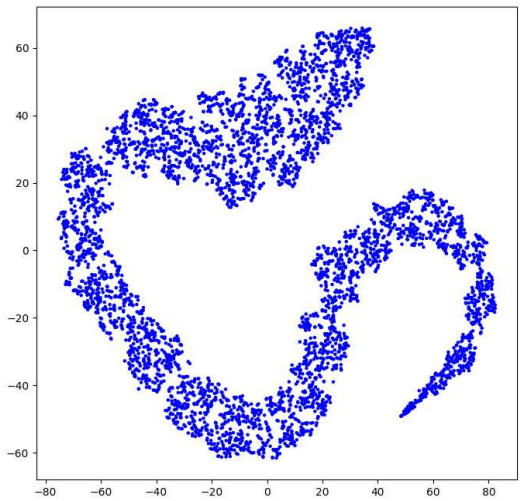

muls

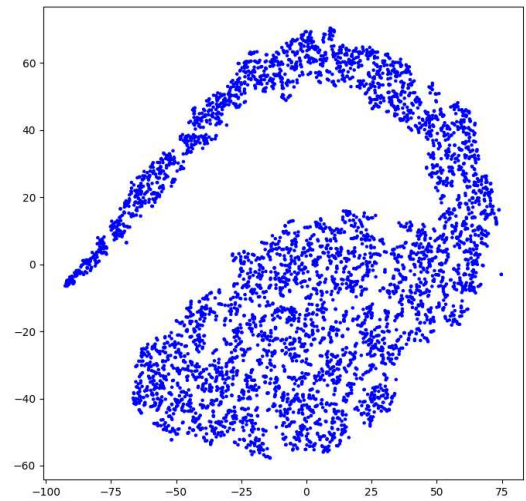

mZje

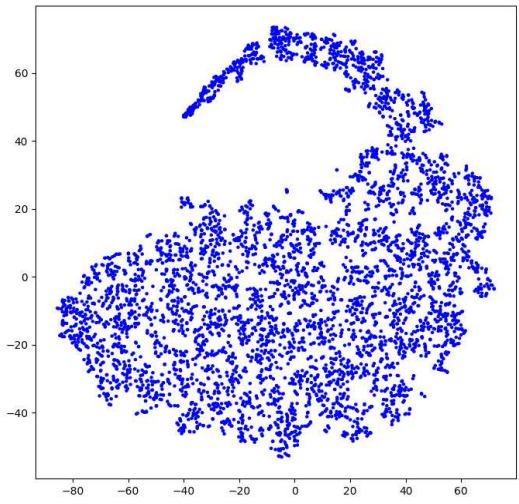

UsSz

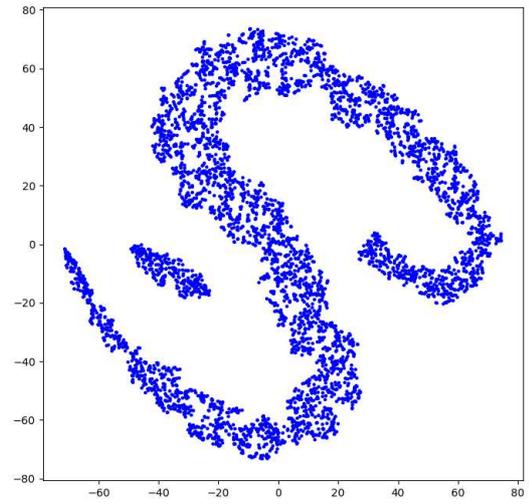

YOh8

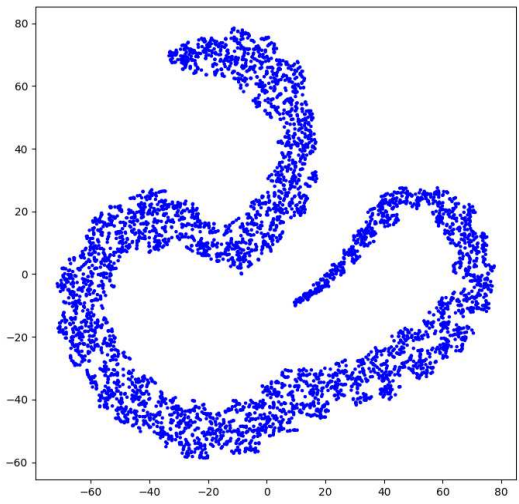

0ncr

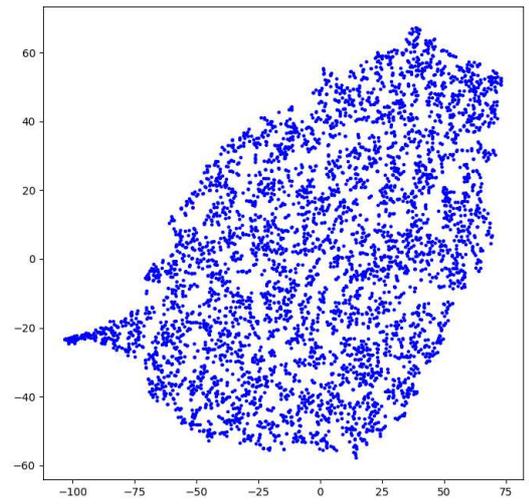

MS6u

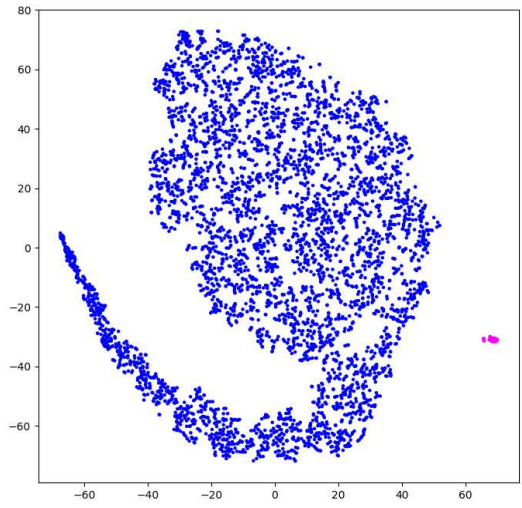

pPpj

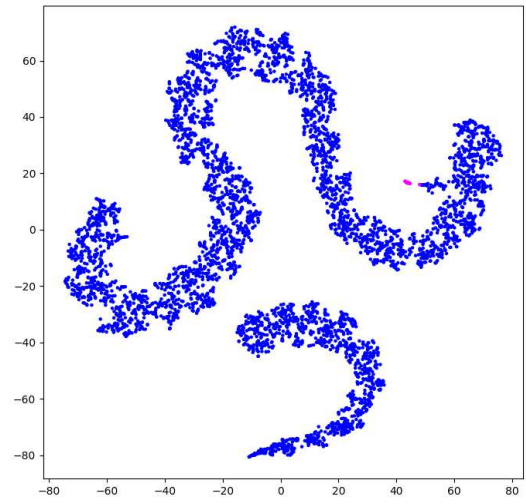

UwK6

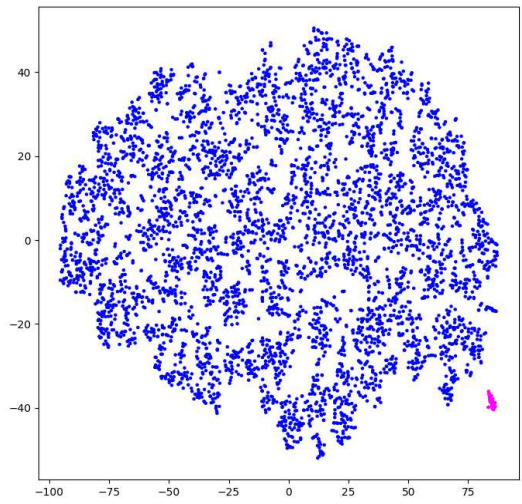

Xii6

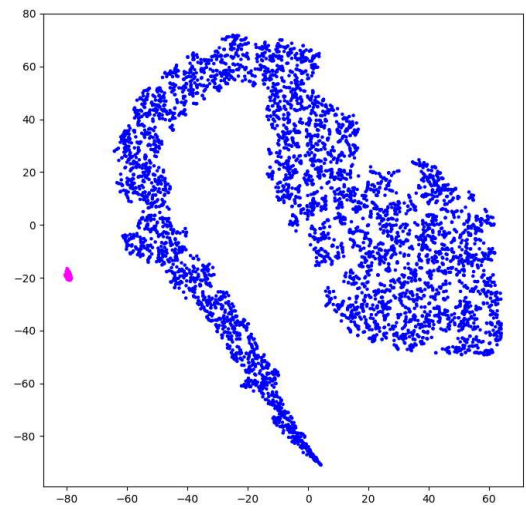

YHLr

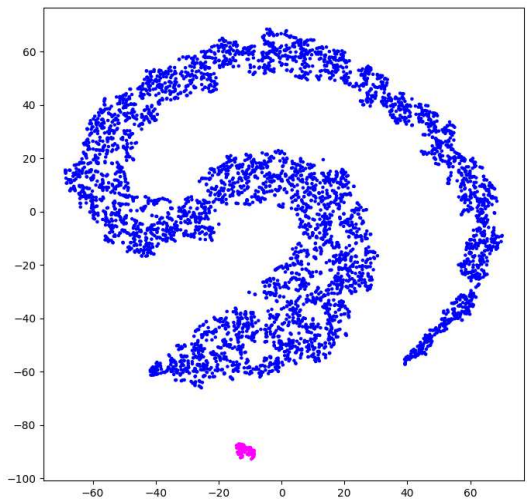

EHED

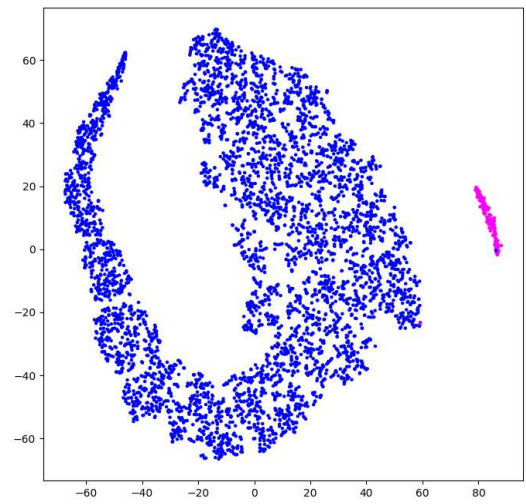

Kn9O

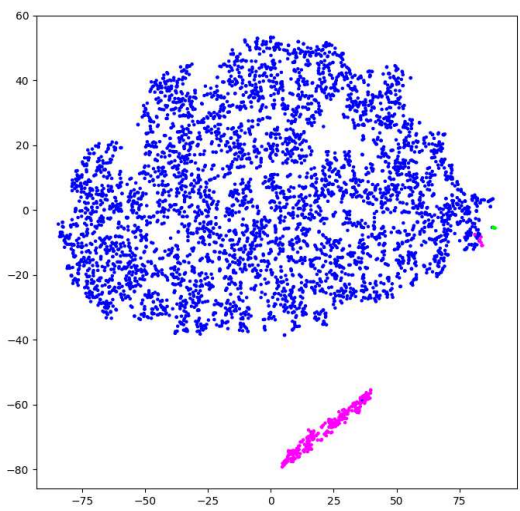

Dr51

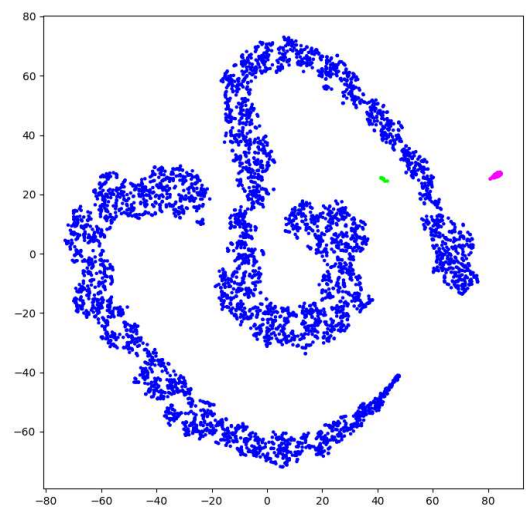

DYYI

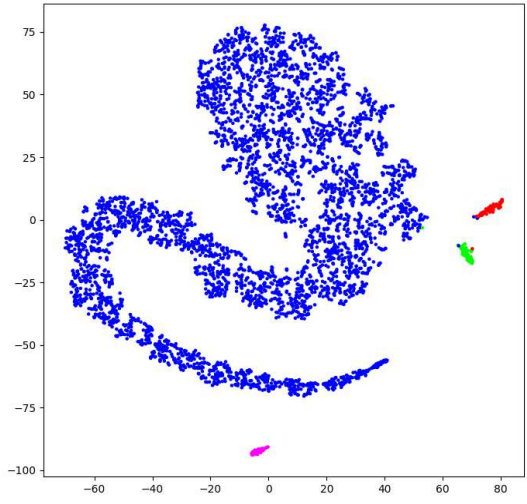

kj2l

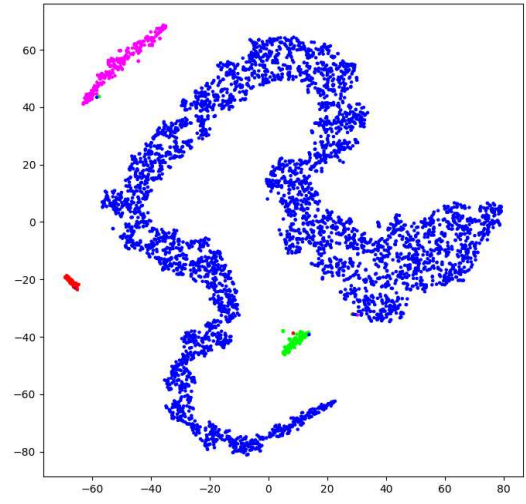

LR2s

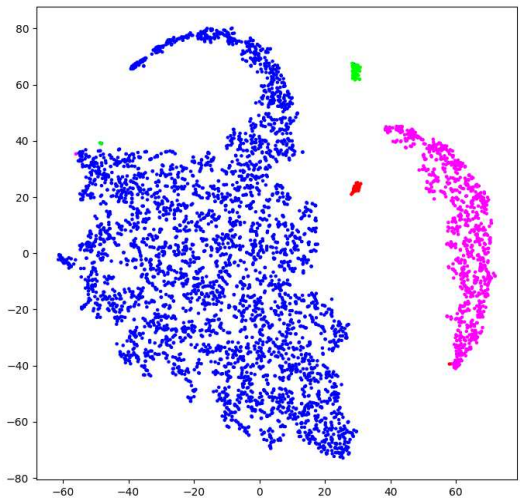

SOZ3

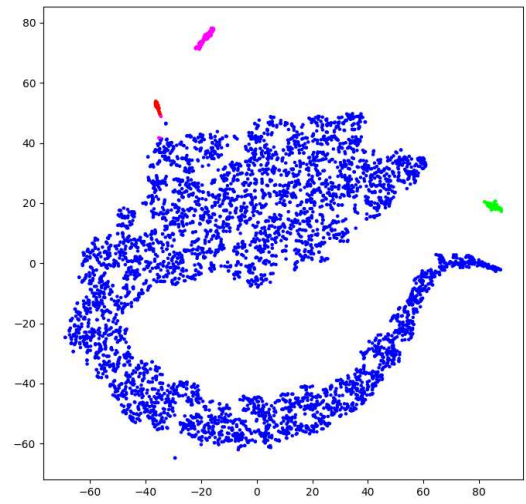

Zpwh

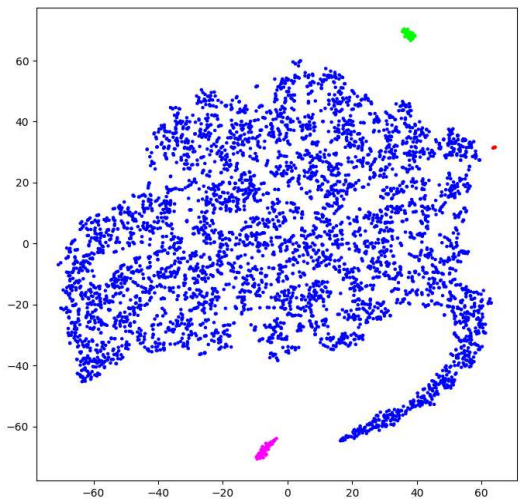

3P0D

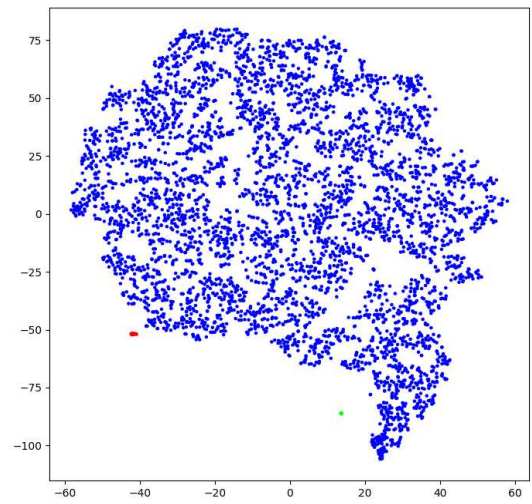

bkx9

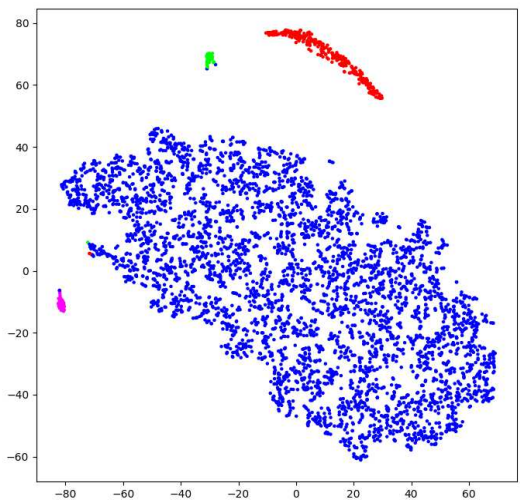

C1Wu

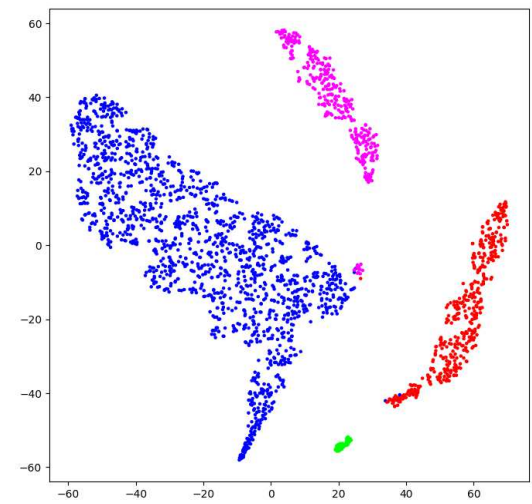

DjrT

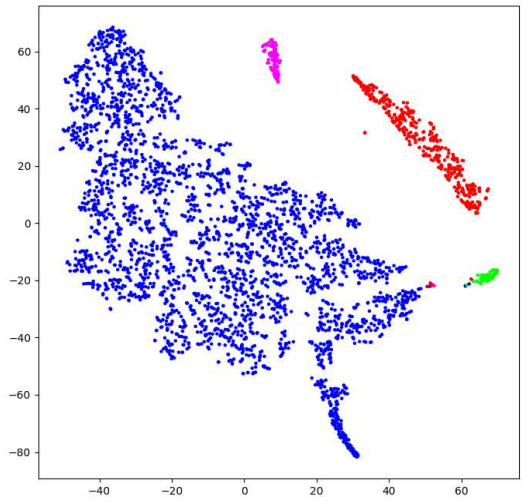

EMcQ

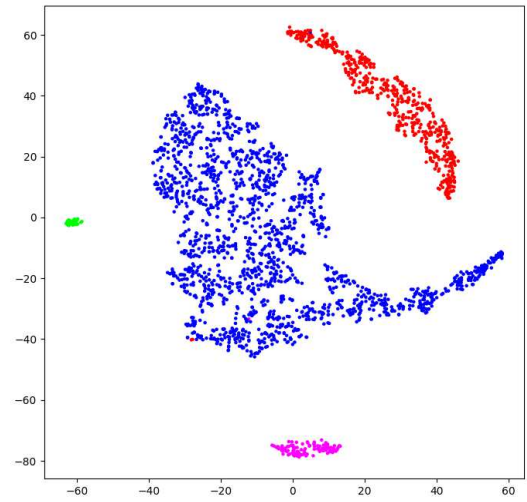

hT38

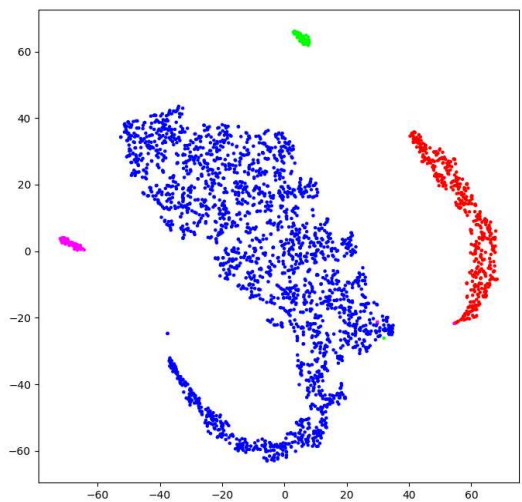

ibbz

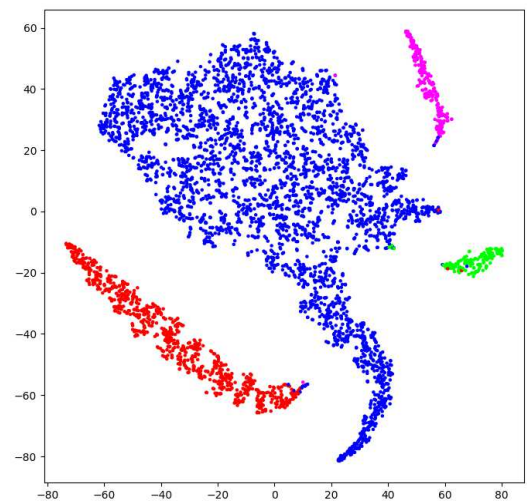

iSqw

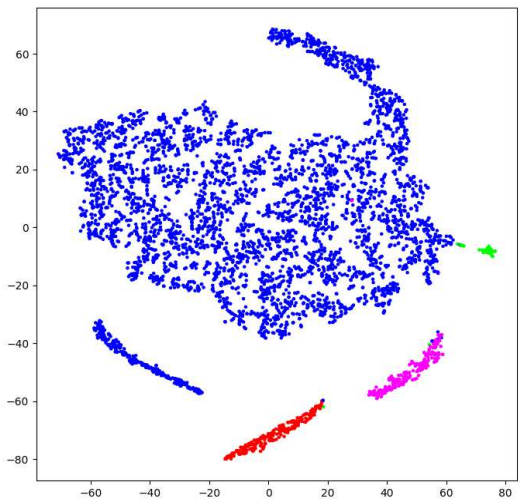

Ivfn

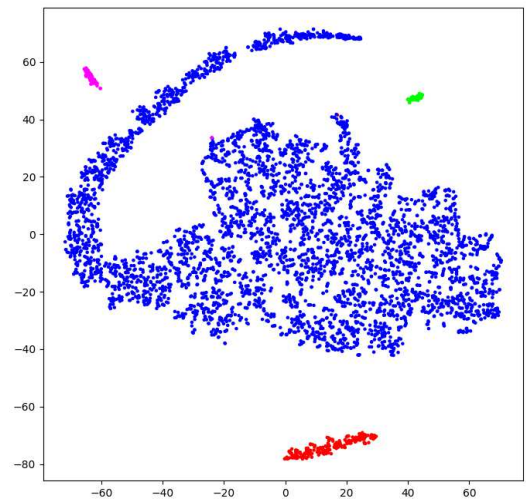

Otq3

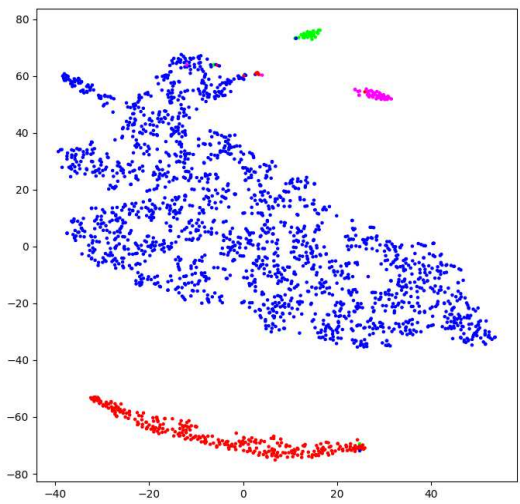

tG6i

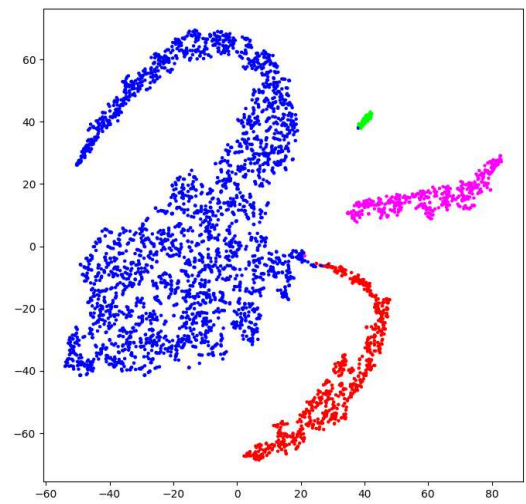

X7s0

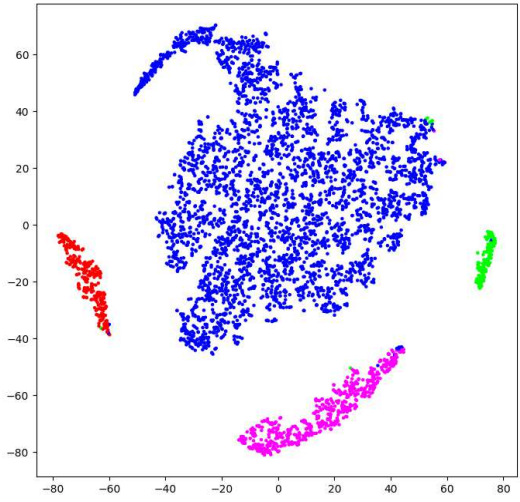

ZYFG

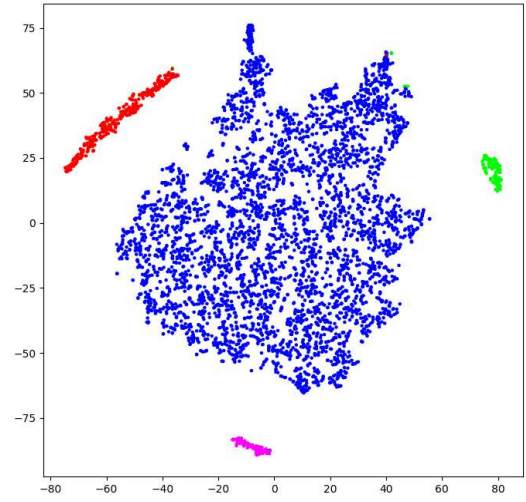

5bSg

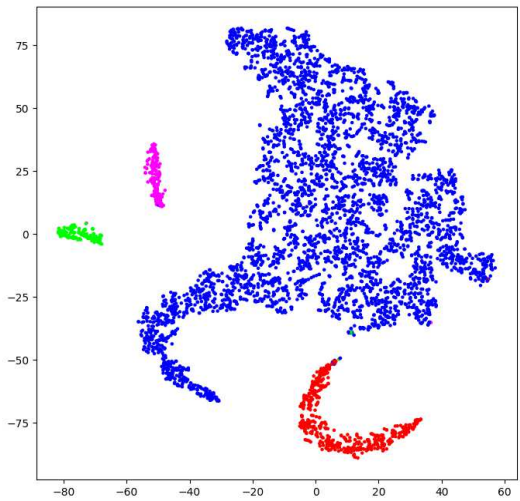

BSvO

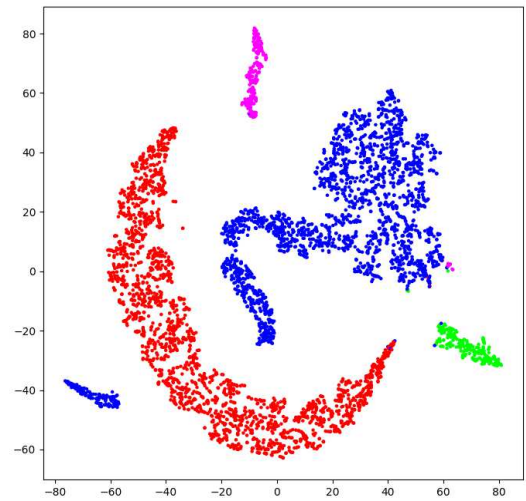

G7PJ

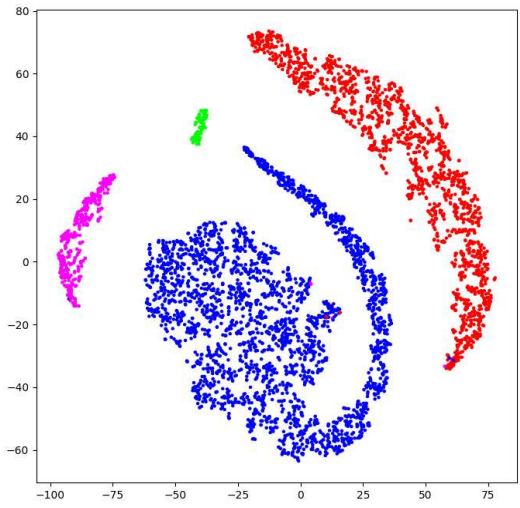

IhpU

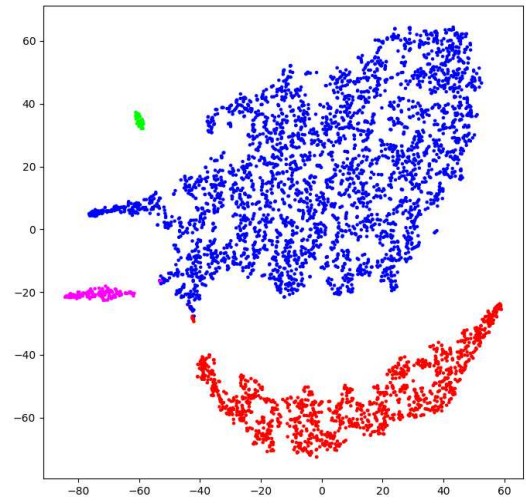

Msy4

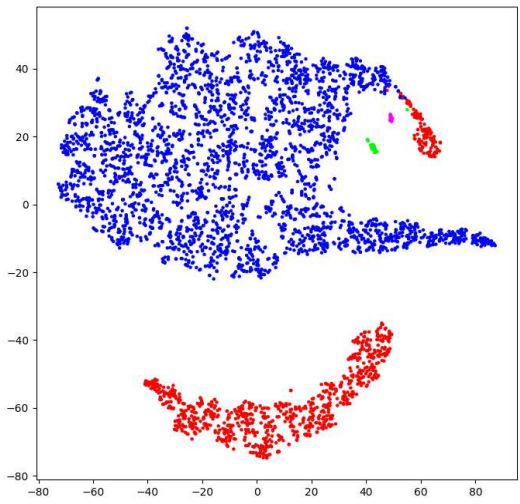

N1nM

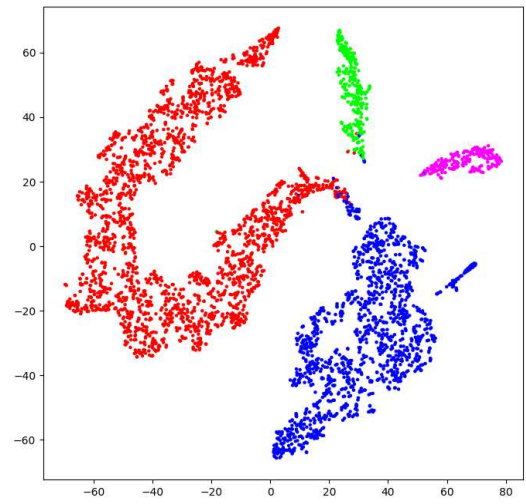

Nzhl

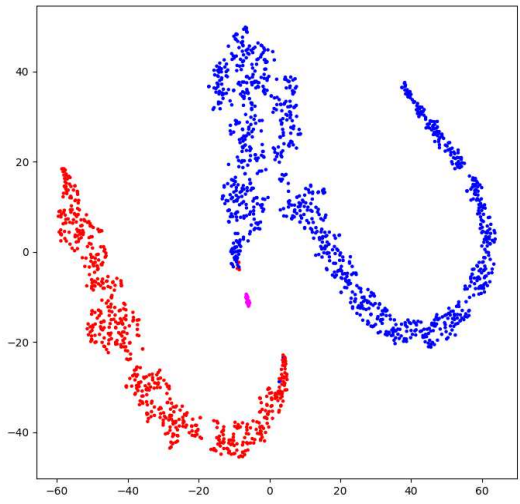 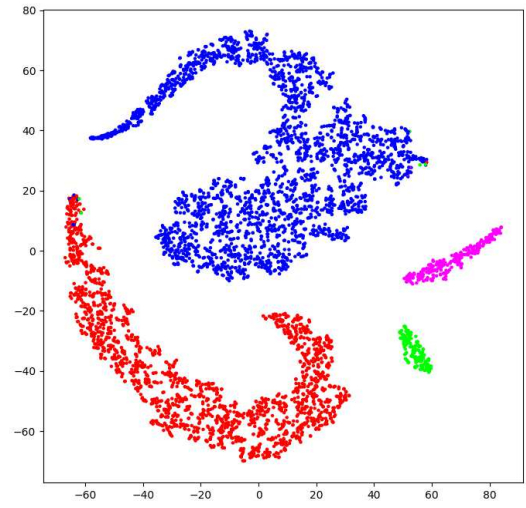

RfL0 svlu

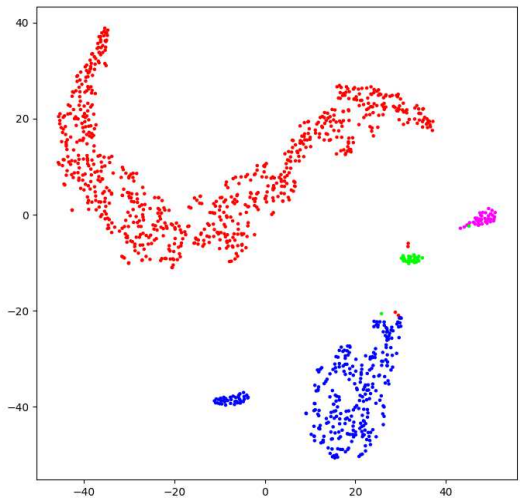 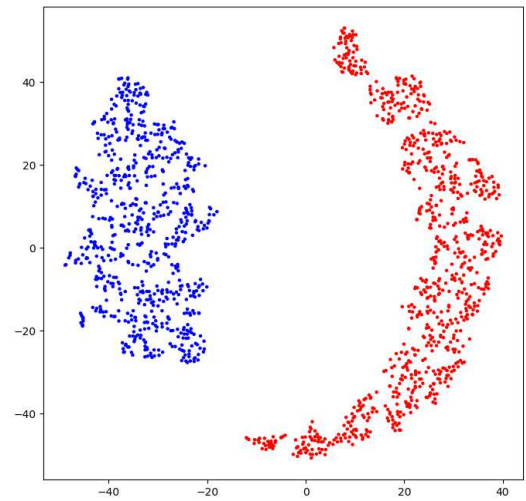

Xg1I

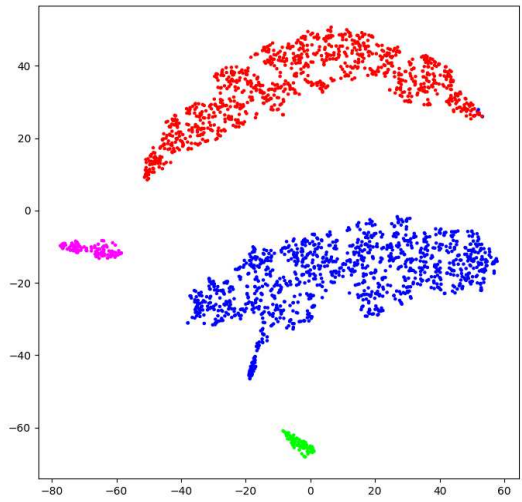

**Supplementary Figure S4.** T-distributed stochastic neighbor embedding (t-SNE) was used to illustrate the 12 patients of the validation dataset and their classification mapped into a 2D space (last layer of the CNN 16s; arbitrary units). Basically, two large clusters corresponding to W and MSE are visible which don't completely separate. MSEc and ED don't form clusters and are not separable from W and MSE. Wakefulness (W): blue; microsleep episodes (MSE): red; microsleep episode candidates (MSEc) green; episodes of drowsiness (ED): magenta. For the convenience we illustrated only every hundredth datapoint (sample). The patient ID is provided at the top of the plot. Please note that these figures only show the internal representation of the data in our specific network. Training data are illustrated in Supplementary Figure S3.

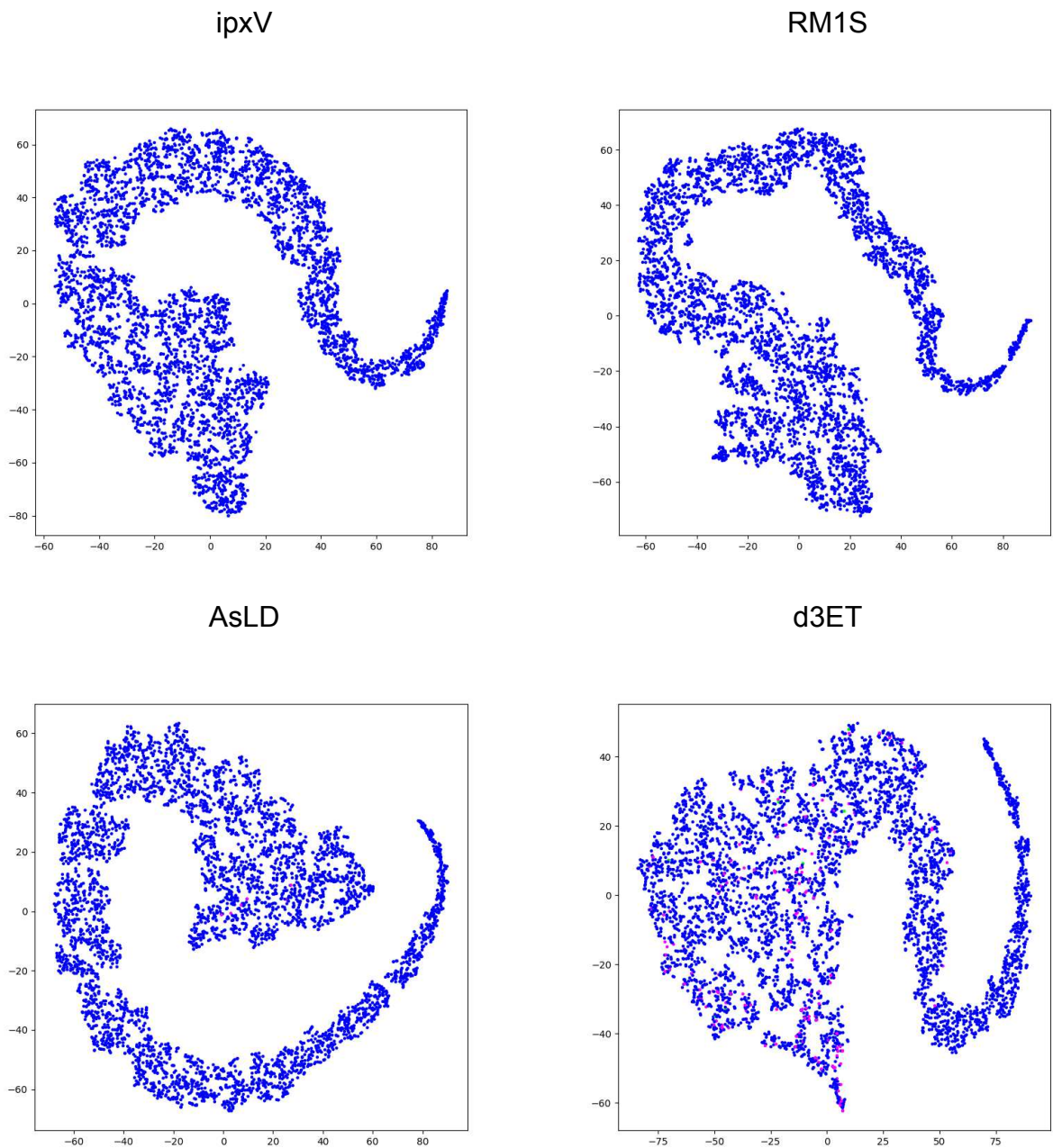

uXdB

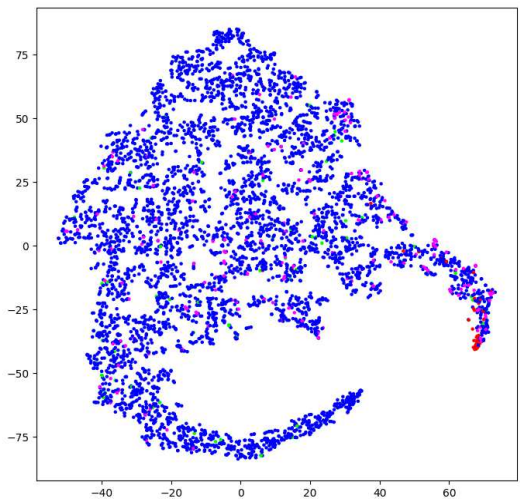

3J4W

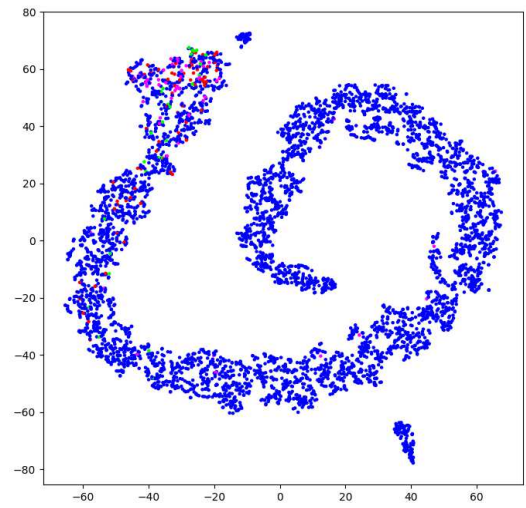

zaca

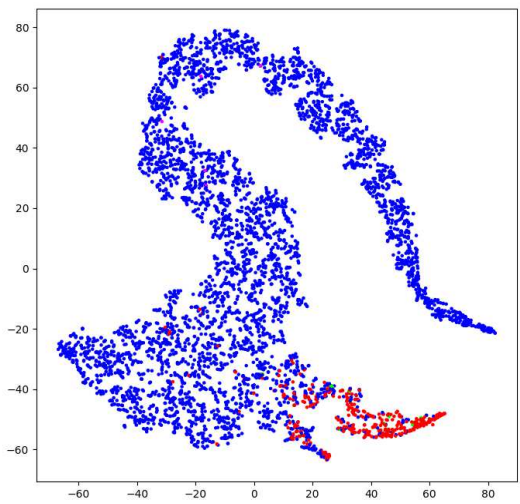

9JQY

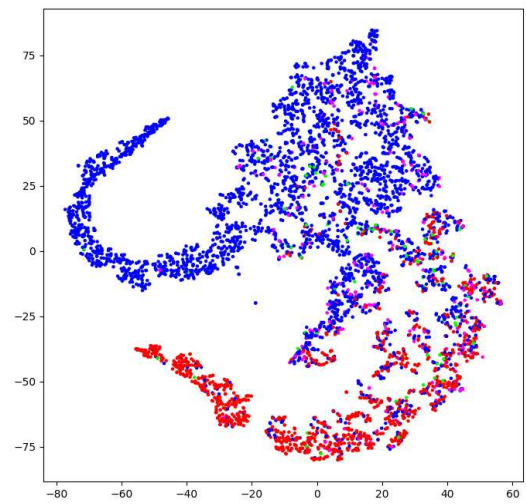

JCpz

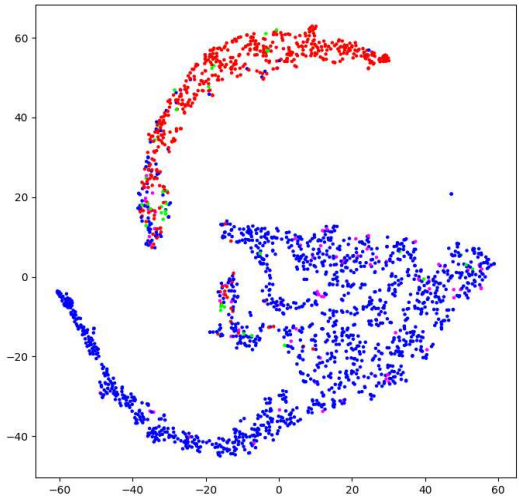

oOMR

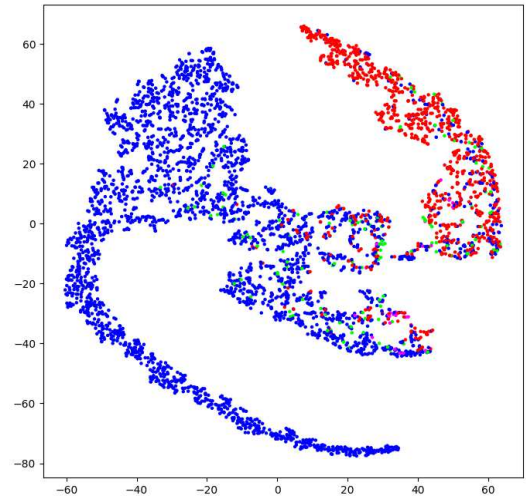

sNMf

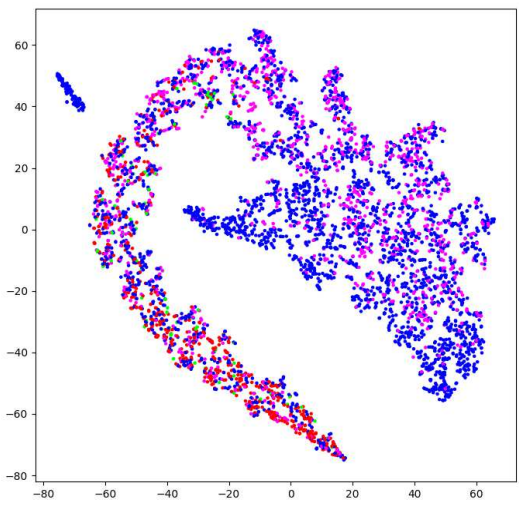

Y5We

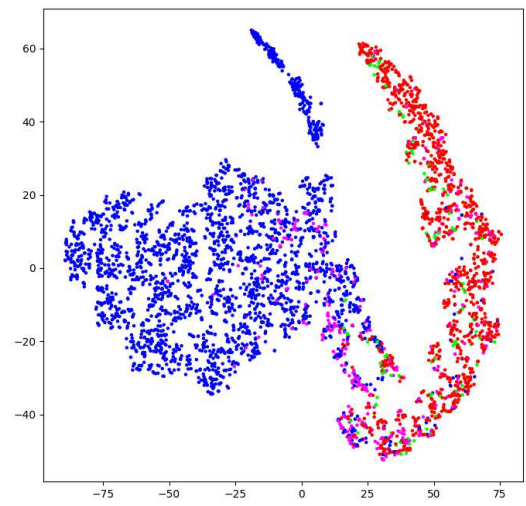